\documentclass[10pt,twocolumn,letterpaper]{article}

\usepackage{iccv}
\usepackage{times}
\usepackage{epsfig}
\usepackage{graphicx}
\usepackage{amsmath}
\usepackage{amssymb}
\usepackage{amsthm}
\usepackage{multirow}
\usepackage{colortbl}
\usepackage{textcomp}

\usepackage{float}
\usepackage{bm}
\usepackage{bbm}
\usepackage{wrapfig}
\usepackage{booktabs}

\usepackage{amsmath,amsfonts,bm}

\def\eqref#1{equation~\ref{#1}}

\def\1{\bm{1}}

\DeclareMathAlphabet{\mathsfit}{\encodingdefault}{\sfdefault}{m}{sl}
\SetMathAlphabet{\mathsfit}{bold}{\encodingdefault}{\sfdefault}{bx}{n}

\newcommand\ours{CALM\xspace}
\newcommand\oursml{$\text{CALM}_\text{ML}$\xspace}
\newcommand\oursem{$\text{CALM}_\text{EM}$\xspace}

\newcommand\maxboxacc{\texttt{MaxBoxAccV2}\xspace}
\newcommand\pxap{\texttt{PxAP}\xspace}
\newcommand\trainweaksup{\texttt{train}\xspace}
\newcommand\trainfullsup{\texttt{val}\xspace}
\newcommand\testfullsup{\texttt{test}\xspace}

\usepackage[pagebackref=true,breaklinks=true,letterpaper=true,colorlinks,bookmarks=false]{hyperref}

\usepackage[pagebackref=true,breaklinks=true,letterpaper=true,colorlinks,bookmarks=false]{hyperref}

\iccvfinalcopy 

\def\iccvPaperID{1308} 

\ificcvfinal\fi

\begin{document}

\title{Keep CALM and Improve Visual Feature Attribution}

\author{
Jae Myung Kim$^{1}\thanks{Equal contribution. Majority of work done at NAVER AI Lab.}$~~~~~~~~~Junsuk Choe$^{2}$\footnotemark[1]~~~~~~~~~Zeynep Akata$^{1,3,4}$~~~~~~~~~Seong Joon Oh$^5$\thanks{Corresponding author.}\\
~\\
\small{
$^1$University of T\"{u}bingen\quad\quad
$^2$Department of Computer Science and Engineering, Sogang University\quad\quad}\\
\small{
$^3$Max Planck Institute for Intelligent Systems\quad\quad
$^4$Max Planck Institute for Informatics\quad\quad
$^5$NAVER AI Lab}
\vspace{-0.5em}
}

\maketitle
\ificcvfinal\thispagestyle{empty}\fi

\begin{abstract}
The class activation mapping, or CAM, has been the cornerstone of feature attribution methods for multiple vision tasks. Its simplicity and effectiveness have led to wide applications in the explanation of visual predictions and weakly-supervised localization tasks. However, CAM has its own shortcomings. The computation of attribution maps relies on ad-hoc calibration steps that are not part of the training computational graph, making it difficult for us to understand the real meaning of the attribution values. In this paper, we improve CAM by explicitly incorporating a latent variable encoding the location of the cue for recognition in the formulation, thereby subsuming the attribution map into the training computational graph. The resulting model, \textbf{\textit{class activation latent mapping}}, or \textbf{\textit{CALM}}, is trained with the expectation-maximization algorithm. Our experiments show that CALM identifies discriminative attributes for image classifiers more accurately than CAM and other visual attribution baselines. CALM also shows performance improvements over prior arts on the weakly-supervised object localization benchmarks. Our code is available at \href{https://github.com/naver-ai/calm}{https://github.com/naver-ai/calm}. 
\vspace{-1em}
\end{abstract}

\section{Introduction}

{Interpretable AI}~\cite{ExplainableAI,InterpretabilityReview1,InterpretabilityReview2,ReviewPaper2021} is becoming an absolute necessity in safety-critical and high-stakes applications of machine learning. Along with good recognition and prediction accuracies, we require models to be able to transparently communicate the inner mechanisms with human users. In visual recognition tasks, researchers have developed various {feature attribution} methods to inspect contributions of individual pixels or visual features towards the final model prediction. Input gradients~\cite{FirstDNNInputGradient,SmoothGrad,FullGrad,LRP,LearnableBackProp,DeepTaylorDecomposition,TCAV} and input perturbation methods~\cite{IntegratedGradients,ZintgrafVisualize,VedaldiMeaingfulPerturbation,RISE,CounterfactualExplanation,Anchors} have been actively researched. 

\begin{figure}
    \centering
    \includegraphics[width=\linewidth]{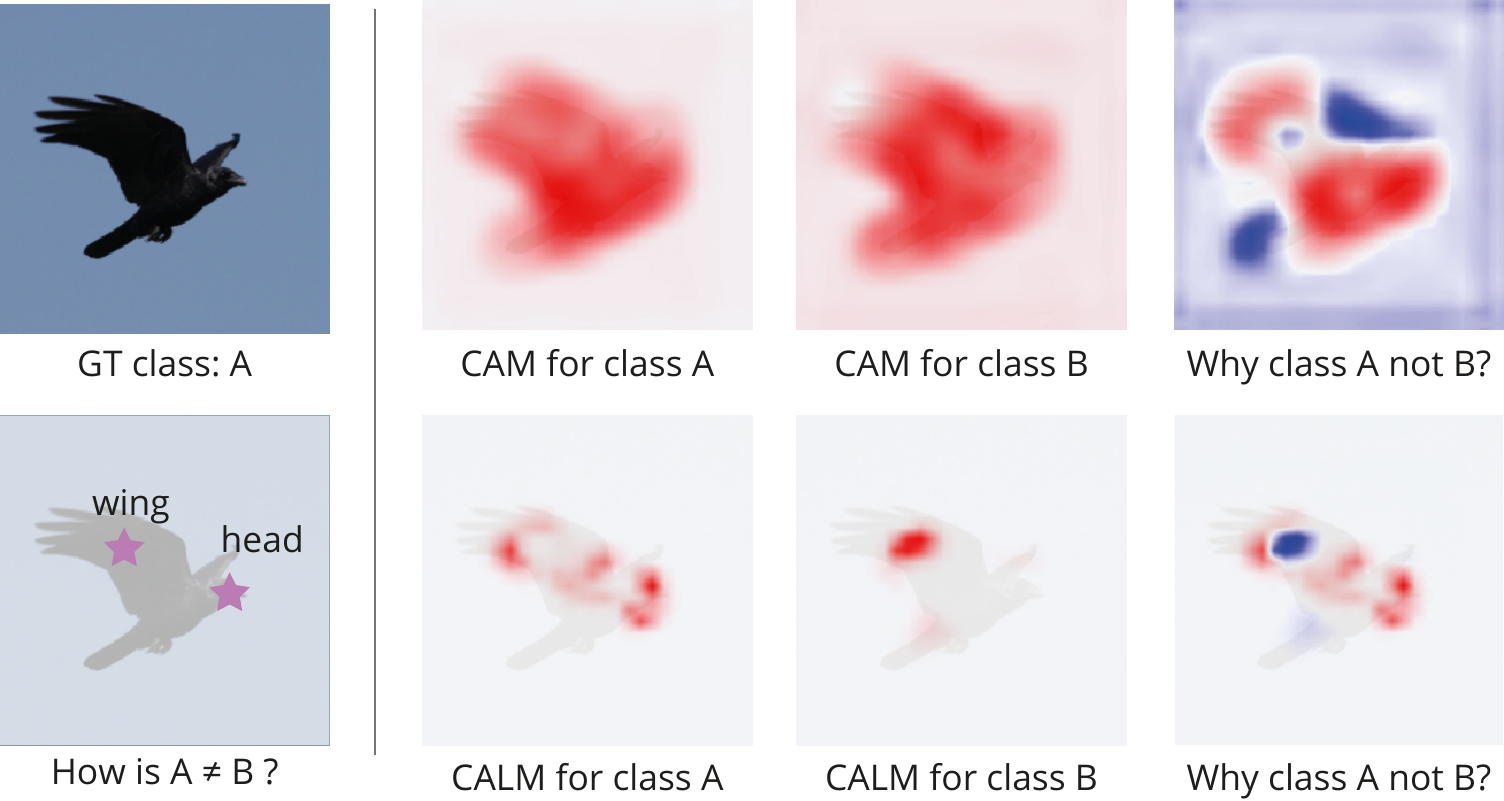}
    \caption{\small\textbf{CAM vs \ours.} \ours is better at locating the actual cues used for the recognition than CAM. Two bird classes A and B only differ in their head and wing attributes. Attributions for class A, B, and their difference are shown. While CAM fails to detect the head and wing, \ours captures them accurately. }
    \vspace{-1em}
    \label{fig:teaser}
\end{figure}

In this paper, we focus on the \textbf{class activation mapping (CAM)}~\cite{CAM} method, which has been the cornerstone of the feature attribution research. CAM starts from the observation that many CNN classifiers make predictions by aggregating location-wise signals. For example, $p(y|x)=\text{softmax}\left(\frac{1}{HW}\sum_{hw} f_{yhw}\right)$ where $f=f(x)$ is the extracted feature map in $\mathbb{R}^{C\times H\times W}$ where $C,H,W$ are the number of classes, height, and width of the feature map, respectively. CAM considers the pre-GAP feature map $f_{yhw}$ as the attribution, after scaling it to the $[0,1]$ range by dropping the negative values and dividing through by the maximum value: $s:=(\max_{hw}f_{hw})^{-1}f^+\in[0,1]^{H\times W}$. Thanks to the algorithmic simplicity and reasonable effectiveness, CAM has been a popular choice as an attribution method with many follow-up variants~\cite{GRADCAM,GradCAMPP,RethinkingCAM,CAMDecompose,ScoreCAM,AblationCAM,XGradCAM}.

Despite its popularity and contributions to the interpretability community, CAM still has its own limitations. What does the attribution map $s$ really mean? We fail to find a reasonable linguistic description because $s$ hardly encodes anything essential in the recognition process. $s$ also violates key minimal requirements, or ``axioms''~\cite{InterpretabilityReview1,IntegratedGradients,XGradCAM}, for an attribution method. For example, its dependence on the pre-softmax values $f$ make it ill-defined: translating $f\mapsto f+c$ yields an identical model because of the translation invariance of softmax, but it changes the attribution map $s$.

We thus introduce a novel attribution method, \textbf{class activation latent mapping (\ours)}. It builds a probabilistic graphical model on the last layers of CNNs, involving three variables: input image $X$, class label $Y\in\{1,\cdots,C\}$, and the location of the cue for recognition $Z\in\{1,\cdots,HW\}$. Since there is no observation for $Z$, we consider latent-variable training algorithms like marginal likelihood (ML) and expectation-maximization (EM). After learning the dependencies, we define the attribution map for image $\widehat{x}$ of class $\widehat{y}$ as $p(\widehat{y},z|\widehat{x})\in[0,1]^{H\times W}$, the joint likelihood of the recognition cue being at $z$ and the class being $\widehat{y}$. \ours has many advantages over CAM. (1) It has a human-understandable probabilistic definition; (2) it satisfies the axiomatic requirements for attribution methods; (3) it is empirically more accurate and useful than CAM.

In our experimental analysis, we study how well \ours localizes the ``correct cues'' for recognizing the given class of interest. The ``correct cues'' for recognition are ill-defined in general, making the evaluation of attribution methods difficult. We build a novel evaluation benchmark on pairs of bird classes in CUB-200-2011~\cite{CUB} where the true cue locations are given by the parts where the attributes for the class pair differ (Figure~\ref{fig:teaser}). Under this benchmark and a widely-used \textit{remove-and-classify} type of benchmark, \ours shows better attribution performances than CAM and other baselines. We also show that \ours advances the state of the art in the weakly-supervised object localization (WSOL) task, where CAM has previously been one of the best~\cite{WSOLEVAL,WSOLEVALJOURNAL}.

In summary, our paper contributes (1) analysis on the lack of interpretability for CAM, (2) a new attribution method \ours that is more interpretable and communicable than CAM, and (3) experimental results on real-world datasets where \ours outperforms CAM in multiple tasks. Our code is available at \href{https://github.com/naver-ai/calm}{https://github.com/naver-ai/calm}.

\section{Related Work}

{Interpretable AI} is a big field. The general aim is to enhance the transparency and trustworthiness of AI systems, but different sub-fields are concerned with different parts of the system and application domains. In this paper, we develop a {visual feature attribution} method for image classifiers based on deep neural networks. It is the task of answering the question: ``how much does each pixel or visual feature contribute towards the model prediction?'' 

\noindent\textbf{Gradient-based attribution.}
Feature attribution with gradients dates back to the pioneering works by Sung~\cite{FirstInputGradient} and Baehrend~\etal~\cite{TheFirstInputGradient}. The first explicit application to CNNs is the work by Simonyan~\etal~\cite{FirstDNNInputGradient}. Input gradients consider local linearization of the model, but it is often not suitable for CNNs because the local behavior hardly encodes the complex mechanisms in CNNs for \eg more global perturbations on the input. Follow-up works have customized the backpropagation algorithm to improve the attribution performances: Guided Backprop~\cite{GuidedBackprop}, LRP~\cite{LRP}, Deep Taylor Decomposition~\cite{DeepTaylorDecomposition}, SmoothGrad~\cite{SmoothGrad}, Full-Gradient~\cite{FullGrad}, and others~\cite{ExcitationBackprop,PatternNet,DeepLIFT,LearnableBackProp,VarGrad,TCAV}. 
We make an empirical comparison against key prior methods in this domain.

\noindent\textbf{Perturbation-based attribution.}
Researchers have developed methods for measuring the model response to non-local perturbations. Integrated Gradients~\cite{IntegratedGradients} measure the path integral of model responses to global input shifts. Another set of methods consider model responses to redacted input parts: sliding windows of an occlusion mask~\cite{ZintgrafVisualize} and random-pixel occlusion masks~\cite{RISE}. Since the occluding patterns introduce artefacts that may mislead attributions, different options for redaction have been considered: ``meaningful perturbations'' like image blurring~\cite{VedaldiMeaingfulPerturbation,VedaldiPerturbation}, inpainting~\cite{ZintgrafVisualize}, and cutting-and-pasting a crop from another image of a different class~\cite{CounterfactualExplanation}.
Some of the key methods above are included as baselines for our experiments.

\noindent\textbf{CAM-based attribution.}
Gradients and perturbations analyze the model by establishing the input-output relationships. Class activation mapping (CAM)~\cite{CAM} takes a different approach. Many CNNs have a global average pooling (GAP) layer towards the end. CAM argues that the pre-GAP features represent the discriminativeness in the image. Related works have considered variants of the last-layer modifications like max pooling~\cite{GMP} and various thresholding strategies~\cite{WELDON,WILDCAT,RethinkingCAM}. GradCAM~\cite{GRADCAM} and GradCAM++~\cite{GradCAMPP} have later expanded the usability of CAM to networks of any last-layer modules by combining the widely-applicable gradient method with CAM.
In this work, we identify issues with CAM and suggest an improvement. 

\noindent\textbf{Self-explainable models.} Above attribution methods provide interpretations of a complex, black-box model in a post-hoc manner. Another paradigm is to design models that are interpretable by design in the first place~\cite{SelfExplainableAI}. There is a trade-off between interpretability and performance~\cite{ExplainableAIReview}; researchers have sought ways to push the boundary on both fronts. One line of work \textit{distills} the complex, performant model into an interpretable surrogate model such as decision trees~\cite{DistillSoftDecisionTree}, sparse linear models~\cite{ExplanationWithAdditiveClassifiers,LIME,SENN}. Other works pursue a \textit{hybrid} approach, where a small interpretable module of a neural network is exposed to humans, while the complex, less interpretable modelling is performed in the rest of the network. ProtoPNet~\cite{LooksLikeThat} trains an interpretable linear map over prototype neural activations. Concept or semantic bottleneck models~\cite{SemanticBottleneckNetwork,ConceptBottleneck} enforces an intermediate layer to explicitly encode semantic concepts. Our work is a \textit{hybrid} self-explainable model based on the interpretable probabilistic treatment of the last layers of CNNs.

\noindent\textbf{Evaluating attribution}
is challenging because of the lack of ground truths. Early works have resorted to qualitative~\cite{CounterfactualExplanation} or human-in-the-loop evaluations~\cite{LIME,GRADCAM,RegulizeInputGrad,InterpretabilityHITL} with limited reproducibility. Wojciech~\etal~\cite{InterpretabilityEvalWojciech} and subsequent works~\cite{PatternNet,RISE,LIME,InterpretabilityBenchmark} have proposed a quantitative measure based on the \textit{remove-and-classify} framework. 
Along a different axis, researchers have focused on the \textit{necessary conditions} for attribution methods. They can be either theoretical properties, referred to as ``axioms''~\cite{IntegratedGradients,XGradCAM} or empirical properties, referred to as ``sanity checks''~\cite{SanityCheckSaliency}.
In our work, we analyze \ours in terms of the axioms (\S\ref{subsec:theoretical-properties-of-calm}) and evaluate it on a remove-and-classify benchmark (\S\ref{subsec:experiments-remove-and-classify}). Additionally, we contribute a new type of evaluation; we compare the attribution map against the \textit{known} ground-truth attributions on a real-world dataset~\cite{CUB} (\S\ref{subsec:experiments-cue-localization}).


\noindent\textbf{Weakly-supervised object localization (WSOL)}
is similar yet different from the feature attribution task. While the latter is focused on detecting the small, class-discriminative cues in the input, the former necessitates the detection of the full object extents. Despite the discrepancy in the objective, CAM has been widely used for both tasks without modification~\cite{WSOLEVAL,WSOLEVALJOURNAL}. We show that \ours, despite being proposed for the attribution task, outperforms CAM on the WSOL task after some additional aggregation operations (\S\ref{subsec:experiments-wsol}).

\section{Class Activation Mapping (CAM)}
\label{sec:CAM}

We cover the background for the class activation mapping (CAM)~\cite{CAM} and analyze its problems. CAM is a feature attribution method for CNN image classifiers. It is applicable to CNNs with the following last layers: 
\begin{align}
    p(y|x)=\text{softmax}\left(\frac{1}{HW}\sum_{hw}f_{yhw}(x)\right)
    \label{eq:cam-training}
\end{align}
where $f(x)$ is feature map from a fully-convolutional network~\cite{FCN} with dimensionality $C\times H\times W$; each channel corresponds to a class-wise feature map. The network is trained with the negative log-likelihood (NLL), also known as cross-entropy, loss. 

At test time, the attribution map is computed by first fetching the pre-GAP feature map $f^{y=\widehat{y}}(x)\in\mathbb{R}^{H\times W}$ for the ground-truth class $\widehat{y}$. CAM then normalizes the feature map $f^{\widehat{y}}$ to the interval $[0,1]$ in either ways:
\begin{align}
    s=
    \begin{cases}
        (f^{\widehat{y}}_{\max})^{-1}\max(0,f^{\widehat{y}}) & \text{max~\cite{CAM}} \\
        (f^{\widehat{y}}_{\max}-f^{\widehat{y}}_{\min})^{-1}(f^{\widehat{y}}-f^{\widehat{y}}_{\min}) & \text{min-max~\cite{GRADCAM} }
    \end{cases}
\label{eq:scoremap-maxnorm}
\end{align}
where $f_{\{\min,\max\}}:=\{\min_{hw},\max_{hw}\}f_{hw}$.

Note that the original CAM paper~\cite{CAM} considers CNNs with an additional linear layer $W\in\mathbb{R}^{C\times L}$ after the pooling (\eg ResNet). It is known that such networks are equivalent to Equation~\ref{eq:cam-training} when we swap the linear and the GAP layers (which are commutative) and treat the linear layer as a convolutional layer with $1\times 1$ kernels~\cite{GRADCAM}. 

\subsection{Limitations of CAM}
\label{subsubsec:what-is-wrong-with-cam}

CAM lacks interpretability. How can we succinctly communicate the attribution value $s_{hw}$ at pixel index $(h,w)$ to others? The best we can come up with is: 
\begin{quote}
    ``The pixel-wise pre-GAP, pre-softmax feature value at $(h,w)$, measured in relative scale within the range of values $[0,A]$ where $A$ is the maximum of the feature values in the entire image.''
\end{quote}
This description is hardly communicable even to experts in image recognition systems, not to mention general users. The difficulty of communication stems from the fact that the attribution scores $s_{hw}$ are not the quantities used by the recognition system; the computational graph for CAM (Equation~\ref{eq:scoremap-maxnorm}) is not part of the training graph (Equation~\ref{eq:cam-training}).

We present the issues with CAM according to the set of axiomatic criteria for attribution methods~\cite{InterpretabilityReview1,IntegratedGradients,XGradCAM}.

\noindent\textbf{Implementation-invariance axiom}~\cite{IntegratedGradients} 
states that two mathematically identical functions, $\phi_1\equiv\phi_2$, shall possess the same attribution maps, regardless of their implementations. CAM violates this axiom. Assume $\phi_1(f):=\text{softmax}(\frac{1}{HW}\sum_{hw}f_{yhw})$ and $\phi_2(f):=\text{softmax}(\frac{1}{HW}\sum_{hw}f_{yhw}+C)$ for some constant $C$. Since the softmax function is translation invariant, $\phi_1\equiv\phi_2$ for any $C$. However, the CAM attribution map for $\phi_2$ varies arbitrarily with $C$: $s=(\max_{hw} f_{hw}+C)^{-1}(f+C)^+$. Min-max normalization is a solution to the problem, but it alone does not let CAM meet other axioms. This observation reveals the inherent limitation of utilizing feature values before softmax (often called ``logits'') as attribution.

\noindent\textbf{Sensitivity axiom}~\cite{IntegratedGradients,XGradCAM}
states that if the function response $\phi(x)$ changes as the result of altering an input value $x_{hw}$ at $(h,w)$, then the corresponding attribution value $s_{hw}$ shall be non-zero. Conversely, if the response is not affected, then $s_{hw}$ shall be zero. CAM fails to satisfy the sensitivity axiom. Depending on the normalization type, CAM assigns zero attributions to $(h,w)$ where $f_{hw}$ is either negative (for max normalization) or smallest (for min-max normalization). However, being assigned a negative or smallest feature value $f_{hw}$ has little connection to the insensitivity of the model to the input value $x_{hw}$.

\noindent\textbf{Completeness (or conservation) axiom}~\cite{IntegratedGradients,XGradCAM,DeepLIFT,LRP}
states that the sum of attributions $\sum_{hw}s_{hw}$ shall add up to
the function output $\phi(x)=p(y|x)$. The completeness criterion is violated by CAM in general because the summation $\sum_{hw}s_{hw}$ for $s$ in Equation~\ref{eq:scoremap-maxnorm} do not match $\phi(x)=p(y|x)$ in Equation~\ref{eq:cam-training}.
In conclusion, CAM fails to satisfy key minimal requirements for an attribution method.

\section{Class Activation Latent Mapping (\ours)}
\label{sec:method}

\begin{figure*}
    \centering
    \includegraphics[width=\linewidth]{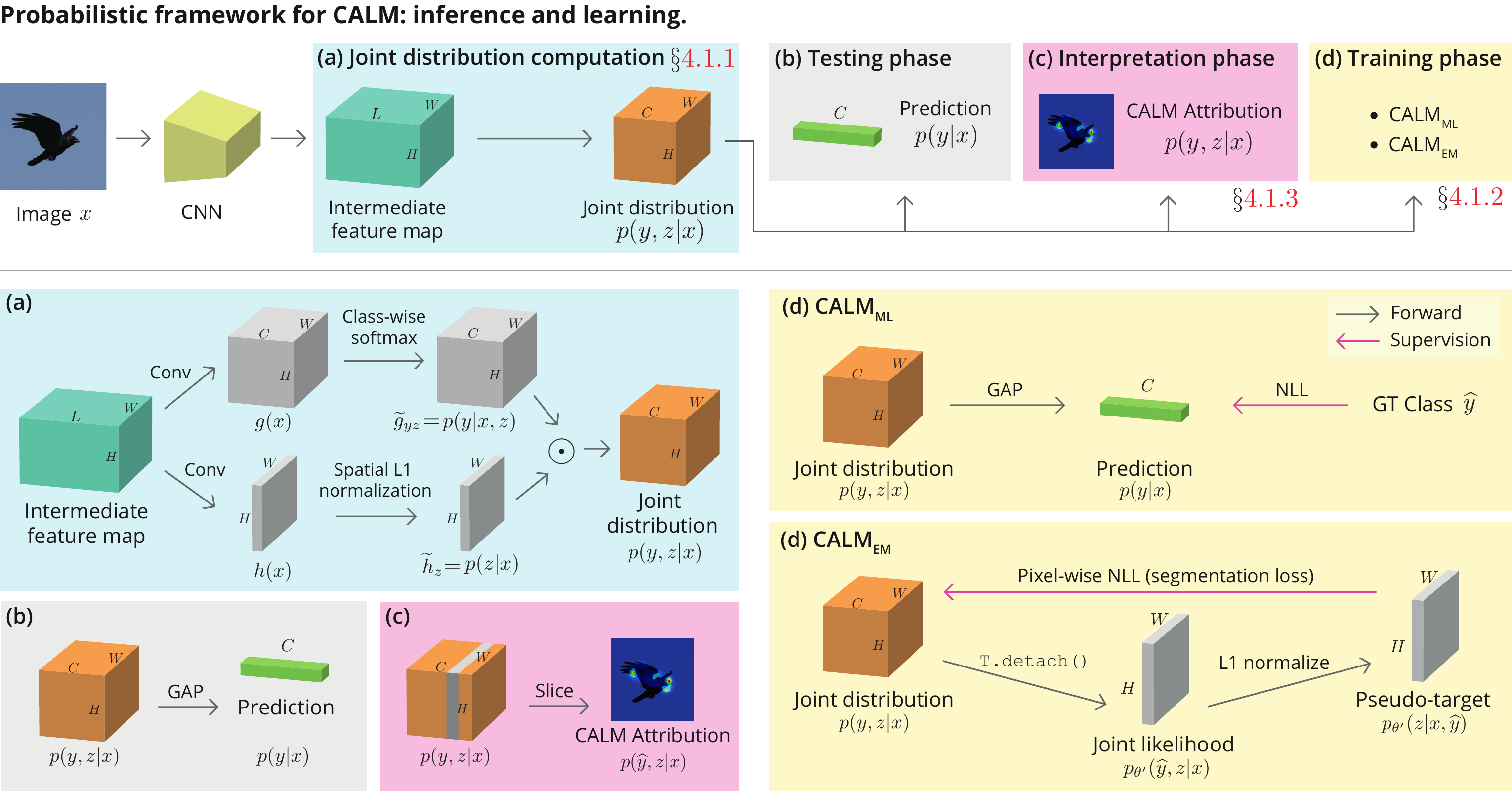}
    \caption{\small\textbf{Main components of \ours.} We show the computational pipeline for \ours during testing, interpretation, and training phases. We zoom into different components. See the relevant sections for more details.}
    \label{fig:method-main}
\end{figure*}

We fix the above issues by introducing a probabilistic learning framework involving \textbf{input image} $X$, \textbf{class label} $Y$, and the \textbf{cue location} $Z$. We set up a probabilistic graphical model and discuss how each component is parametrized with a CNN. We then introduce learning algorithms to account for the unobserved latent variable $Z$. An overview of our method is provided in Figure~\ref{fig:method-main}.

\subsection{Probabilistic inference with latent $Z$}
\label{subsubsec:probabilistic-inference}

We define $Z$ as the location index $(h,w)$ of the cue for recognizing the image $X$ as the class $Y$. Our aim is to let the model explicitly depend its prediction on the features corresponding to the location $Z$ and later on use the distribution of possible cue locations $Z$ as the attribution provided by the model. $Z$ is a random variable over indices $(h,w)$; for simplicity, we use the integer indices $Z\in\{1,\cdots,HW\}$.

\setlength{\columnsep}{1em}%
\begin{wrapfigure}{l}{0.3\linewidth}
    \centering
    \includegraphics[width=.9\linewidth]{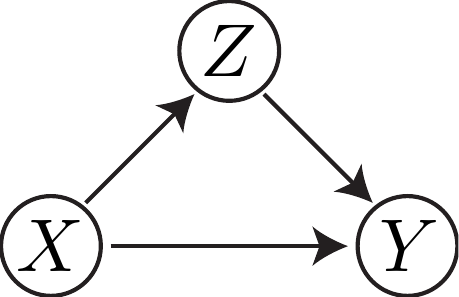}
    \vspace{-1em}
\end{wrapfigure}
Without loss of generality, we factorize $p(x,y,z)$ as $p(y,z|x)p(x)=p(y|x,z)p(z|x)p(x)$ (graph on the left). The recognition task is then performed via $p(y|x)=\sum_z p(y,z|x)$.

\subsubsection{Representing joint distribution with CNNs.}

We factorize the joint distribution $p(y,z|x)$ into $p(y,z|x)=p(y|x,z)p(z|x)$ and parametrize $p(y|x,z)$ and $p(z|x)$ as two convolutional branches of a CNN trunk (Figure~\ref{fig:method-main}a).
Since $Y\in\{1,\cdots,C\}$ and $Z\in\{1,\cdots,HW\}$, we represent $p(y|x,z)$ as a CNN branch $g(x)$ with output dimensionality $C\times HW$. Likewise, we represent $p(z|x)$ with a CNN branch $h(x)$ with output dimensionality $HW$. 
To make sure that the outputs of the two branches are proper distributions, we normalize the outputs with softmax for $g$ and $\ell_1$ normalization followed by the softplus for $h$. We broadcast $h$ to all class indices $Y$ and multiply it element-wise with $g$ to get $p(y,z|x)$  (Figure~\ref{fig:method-main}a).

\subsubsection{Training algorithms}
\label{sec:training_algorithms}

Training a latent variable model is challenging because of the unobserved variable $Z$. We consider two methods for training such a model: (1) marginal likelihood (ML)~\cite{Book1} and (2) expectation-maximization (EM)~\cite{EM}. 

\paragraph{\oursml} 
directly minimizes the marginal likelihood
\begin{align}
    -\log p_\theta({y}|{x}) &=
    -\log \sum_z p_\theta({y}|{x},z) p_\theta(z|{x}) \\
    &= -\log \sum_z {g}_{{y}z}
    \cdot {h}_z.
    \label{eq:calm-ml-implementation}
\end{align}
which is tractable for the discrete $Z$. See Figure~\ref{fig:method-main}d.

\paragraph{\oursem} 
is based on the EM algorithm that generates pseudo-targets for $Z$ to supervise the joint likelihood $p({y},z|{x})$. The EM algorithm introduces two running copies of the parameter set: $\theta$ and $\theta^\prime$. The first signifies the model of interest, while the latter often refers to a slowly updated parameter used for generating the pseudo-targets for $Z$. The learning objective is
\begin{align}
    -\log p_\theta({y}|{x})
    &\leq-\sum_z p_{\theta^\prime}(z|{x},{y}) \log p_\theta({y},z|{x}) \\
    &=-\sum_z \frac{{g}^\prime_{{y}z}\cdot{h}^\prime_z}{\sum_{l}{g}^\prime_{{y}l}\cdot{h}^\prime_l} \log \left({g}_{{y}z}\cdot{h}_z\right) 
    \label{eq:calm-em-implementation}
\end{align}
where $g^\prime$ and $h^\prime$ denote the parametrization with $\theta^\prime$. 
Note that Equation~\ref{eq:calm-em-implementation} is the pixel-wise negative log likelihood, the loss function for semantic segmentation networks~\cite{DeepLab}. One may interpret the objective as self-supervising the pixel($z$)-wise predictions $p(y,z|x)$ with its own estimation of the cue location $z$ for the true class ${y}$: $p_{\theta^\prime}(z|x,{y})$. 
In practice, we use the current-iteration model parameter $\theta=\theta^\prime$ to generate the pseudo-target for $Z$. See Figure~\ref{fig:method-main}d for an overview of the process. Even with $\theta=\theta^\prime$, we need to apply \texttt{T.detach()} to block the gradient flow through the pseudo-target $p_{\theta^\prime}(z|x,{y})$, as required by Equation~\ref{eq:calm-em-implementation}.

A similar framework appears in the weakly-supervised semantic segmentation task. Papandreou~\etal~\cite{WSSS-EM} have generated pseudo-target label maps to train a segmentation network. \oursem is different because our location-encoding latent $Z$ takes integer values, while their $Z$ takes values in the space of all binary masks; our formulation admits an exact computation of Equation~\ref{eq:calm-em-implementation}, while theirs require an additional approximation step.


\subsubsection{Inferring feature attributions}
\label{subsubsec:attribution-map-for-calm}

Unlike CAM, our probabilistic formulation enables principled computation of the attribution map as part of the probabilistic inference on $p(y,z|x)$. $Z$ is explicitly defined as the location of the cue for recognition. For \ours, the \textbf{attribution score} $s_z$ for location $z$ is naturally defined as the joint likelihood given the ground-truth class $\widehat{y}$
\begin{align}
    s_z:=p(\widehat{y},z|{x}),
    \label{eq:calm-attribution-map}
\end{align}
or in human language,
\begin{quote}
    ``The probability that the cue for recognition was at $z$ and the ground truth class $\widehat{y}$ was correctly predicted for the image ${x}$.''
\end{quote}
Note that the definition is far more communicable than the one for CAM in \S\ref{subsubsec:what-is-wrong-with-cam}. See Figure~\ref{fig:method-main}c for visualization.

\begin{figure}
    \centering
    \includegraphics[width=\linewidth]{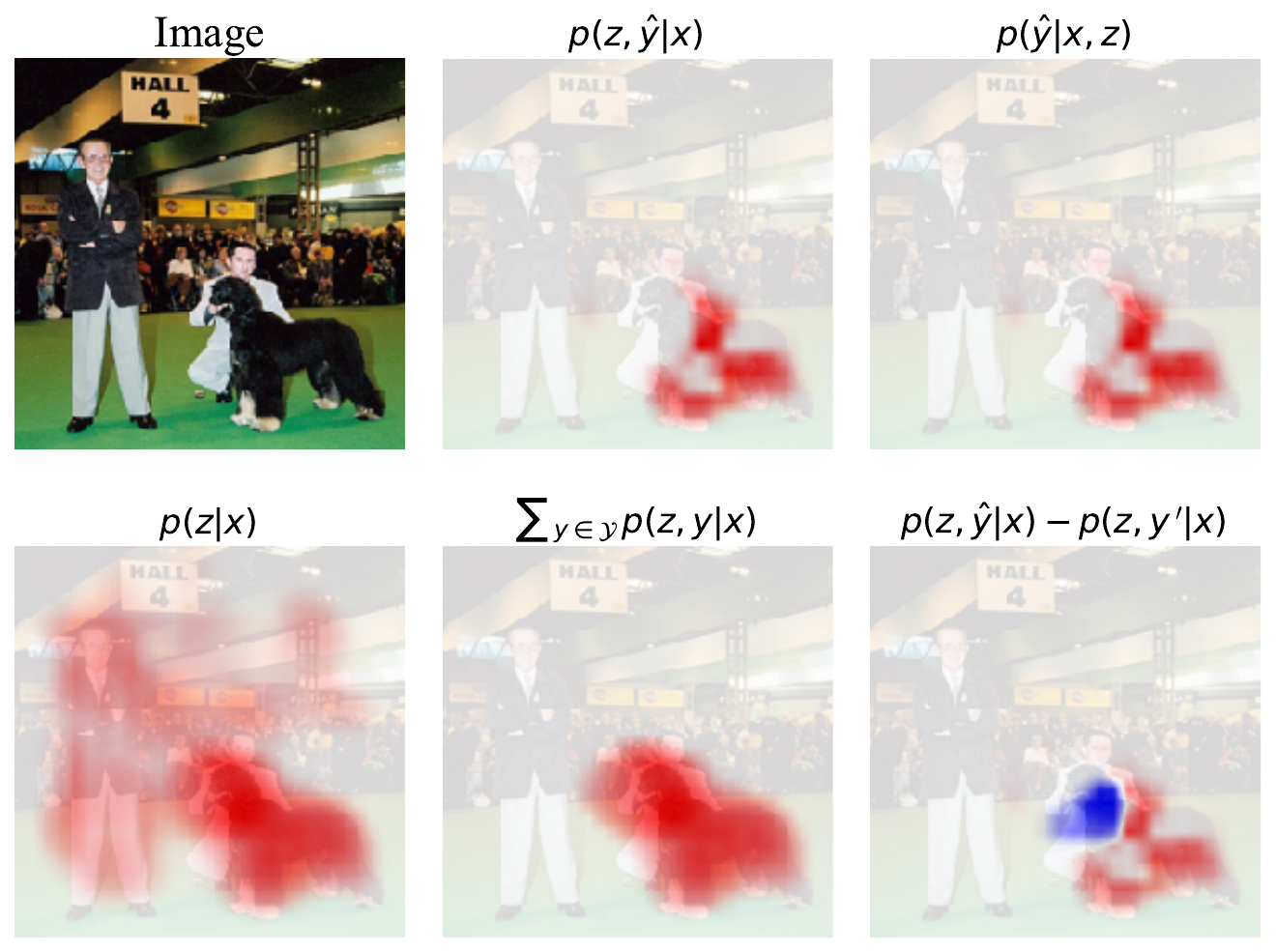}
    \caption{\small\textbf{Various attribution maps by \ours on ImageNet.} GT class is ``Afghan hound''. For the subset attribution, the classes $\mathcal{Y}$ correspond to all species of dogs in ImageNet. For the counterfactual attribution, the alternative class $y^\prime$ is ``Gazelle hound''. }
    \label{fig:attribution-types}
\end{figure}

Apart from the attribution map, one may compute additional interesting quantities. We show examples in Figure~\ref{fig:attribution-types}. {Treating $z$ as a free variable, the \textbf{conditional attribution} $p(y|x,z)$ is explained as the likelihood of the cue being at position $z$, given the prediction for image $x$ as $y$.} The \textbf{saliency} $p(z|x)$ encodes the likely location of any cue for recognizing classes $y\in\{1,\cdots,C\}$ in image $x$. It is the sum over all attribution maps for classes $y$: $p(z|x)=\sum_y p(y,z|x)$. One may also compute the partial sum for classes $y\in\mathcal{Y}$ to obtain the \textbf{subset attribution} to highlight specific image regions of interest $p(z,\mathcal{Y}|x):=\sum_{y\in \mathcal{Y}} p(z,y|x)$. Above quantities are later utilized for the weakly-supervised object localization (WSOL) task in \S\ref{subsec:experiments-wsol}. 
It is also possible to reason why the class label for input $x$ is $\widehat{y}$ instead of $y^\prime$ by computing the \textbf{counterfactual attribution} $p(\widehat{y},z|x)-p(y^\prime,z|x)$. Such counterfactual reasoning will be used in our analysis in \S\ref{subsec:experiments-cue-localization}. 

\begin{figure*}
    \centering
    \includegraphics[width=\linewidth]{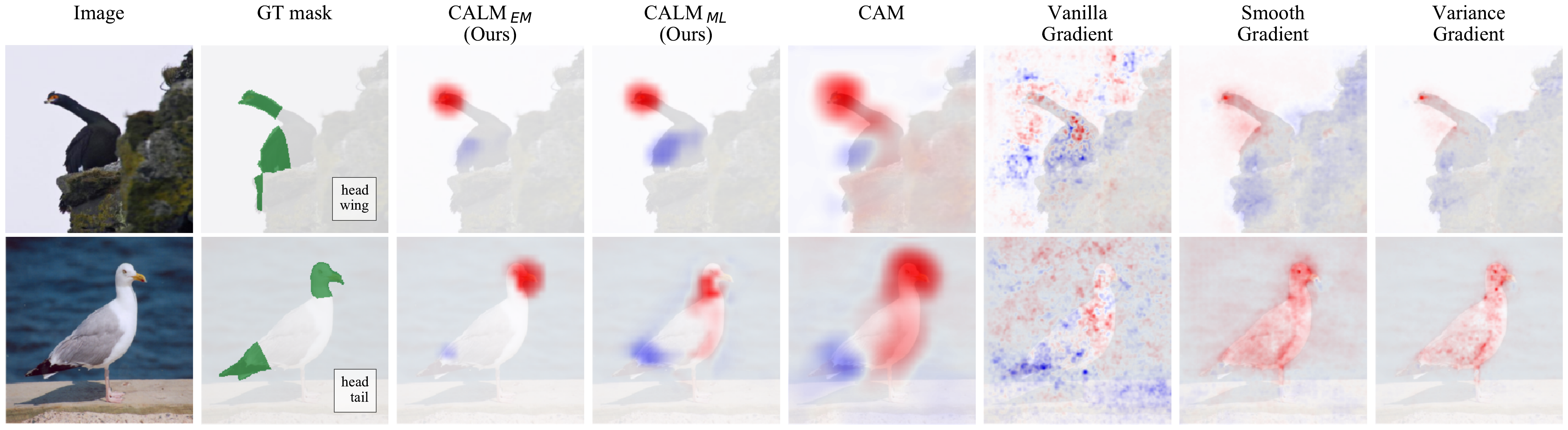}
    \caption{\small\textbf{Qualitative results on CUB.} We compare the counterfactual attributions from \ours and baseline methods against the GT attribution mask. The GT mask indicates the bird parts where the attributes for the class pair (A,B) differ. The counterfactual attributions denote the difference between the maps for classes A and B: $s^\text{A}-s^\text{B}$. Red: positive values. Blue: negative values.}
    \label{fig:interpretability-qualitative-cub}
\end{figure*}

\subsection{Theoretical properties of \ours}
\label{subsec:theoretical-properties-of-calm}

Now we revisit the axioms for attribution methods that CAM fails to fulfill (\S\ref{subsubsec:what-is-wrong-with-cam}). \textbf{Implementation-invariance axiom} is satisfied by \ours because the attribution map $s:=p(\widehat{y},z|x)$ is a mathematical object in the probabilistic graphical model. \ours attribution also does not depend on the fragile logit values. The \textbf{completeness axiom} trivially follows from \ours because the final prediction $p(y|x)$ is the sum of attribution values $p(y,z|x)$ over $z$. Likewise, the \textbf{sensitivity axiom} follows trivially from the fact that $p(y,z|x)>0$ if and only if it contributes towards the sum $p(y|x)=\sum_z p(y,z|x)$.

The superior interpretability of \ours comes with a cost to pay. It alters the formulation of the usual structure for CNN classifiers where the loss function has the structure ``$\text{NLL}\circ\text{SoftMax}\circ\text{Pool}$'' on the feature map $f$ into the one with the structure ``$\text{Pool}\circ\text{NLL}\circ\text{SoftMax}$'' on $f$. Compared to the former, \ours gain additional interpretability by making the last layer of the network as simple as a sum over the pixel-wise experts $p(y,z|x)$. The reduced complexity in turn increases the representational burden for the earlier layers $f(x)$ and induces a drop in the classification accuracies (\S\ref{subsec:experiments-remove-and-classify}). 

The interpretability-performance trade-off is unavoidable~\cite{ExplainableAIReview}. Therefore, it benefits users to provide a diverse array of models with different degrees of interpretability and performance~\cite{StopExplainingBB}. Our work contributes to this diversity of the ecosystem of models.

\section{Experiments}

We present experimental results for \ours. We present two experimental analyses on attribution qualities: evaluation with respect to estimated ground truth attributions on CUB (\S\ref{subsec:experiments-cue-localization}) and the remove-and-classify results on three image classification datasets (\S\ref{subsec:experiments-remove-and-classify}). We then show results on the weakly-supervised object localization task in \S\ref{subsec:experiments-wsol}. Naver Smart Machine Learning (NSML) platform~\cite{NSML} has been used in the experiments.

\subsection{Implementation details and baselines}
\label{subsec:baseline-attribution-methods}

\noindent\textbf{Backbone.} \ours is backbone-agnostic, as long as it is fully convolutional. We use the ResNet50 as the feature extractor $f$ unless stated otherwise. As discussed in \S\ref{sec:CAM}, we move the final linear layer before the global average pooling layer as a convolutional layer with $1\times 1$ kernels.

\noindent\textbf{Datasets.} Our experiments are built on three real-world image classification datasets: CUB-200-2011~\cite{CUB}, a subset of OpenImages~\cite{OpenImagesV5}, and ImageNet1K~\cite{ImageNet}. CUB is a fine-grained bird classification dataset with 200 bird classes. We use a subset of OpenImages curated by \cite{WSOLEVALJOURNAL} that consists of 100 coarse-grained everyday objects. ImageNet1K has 1000 classes with mixed granularity, ranging from 116 fine-grained dog species to coarse-grained objects and concepts.

\noindent\textbf{Pretraining.}
We use the ImageNet pre-trained weights for $f$. The two convolutional layers for computing $p(y,z|x)$ in Figure~\ref{fig:method-main}a are trained from scratch.

\noindent\textbf{Attribution maps.}
The attribution maps for CAM and \ours are scaled up to the original image size via bilinear interpolation. For gradient-based baseline attribution methods, we apply Gaussian blurring and min-max normalization, following \cite{WSOLEVALJOURNAL}.

\noindent\textbf{Other training details} are in the Supplementary Materials.

\subsection{Cue localization results}
\label{subsec:experiments-cue-localization}

\begin{table}[]
    \setlength{\tabcolsep}{.4em}
    \centering
    \small
    \begin{tabular}{ccccccc} 
    \#\ignorespaces part differences && 1             & 2             & 3             &&  \\ \cline{1-1} \cline{3-5}
    \#\ignorespaces class pairs      && 31            & 64            & 96            &&     mean                  \\
    \cline{1-1} \cline{3-5} \cline{7-7}
    \vspace{-1em} & \\
    \cline{1-1} \cline{3-5} \cline{7-7}
    \vspace{-1em} & \\
    Vanilla Gradient \cite{FirstDNNInputGradient}           && 10.0          & 13.7          & 15.3          && 13.9                  \\
    Integrated Gradient \cite{IntegratedGradients}     && 12.0          & 15.1          & 17.3	        && 15.7                  \\
    Smooth Gradient \cite{SmoothGrad}          && 11.8          & 15.5          & 18.6          && 16.5                  \\
    Variance Gradient \cite{VarGrad}            && 16.7          & 21.1          & 23.1	        && 21.4                  \\
    \cline{1-1} \cline{3-5} \cline{7-7}
    \vspace{-1em} & \\
    CAM \cite{CAM}                   && 24.1          & 28.3          & 32.2          && 29.6                  \\
    \oursml (Ours)           && 23.6        &  26.7         & 28.8     && 27.3 \\ 
    \oursem (Ours)           && \textbf{30.4} & \textbf{33.3} & \textbf{36.3} && \textbf{34.3} \\ 
    \cline{1-1} \cline{3-5} \cline{7-7}    
    \end{tabular}
    \vspace{.5em}
    \caption{\small\textbf{Attribution evaluation on CUB.} We use the estimated GT attribution masks (\S\ref{subsec:experiments-cue-localization}) to measure the performances of attribution methods. Mean pixel-wise average precision (mPxAP) values are reported. See Figure~\ref{fig:interpretability-qualitative-cub} for the setup and examples.}
    \label{tab:interpretability-part-retrieval}
    \vspace{-1em}
\end{table}

The difficulty of attribution evaluation comes from the fact that it is difficult to obtain the ground truth cue locations $\widehat{z}$. We propose a way to estimate the true cue location using the rich attribute and part annotations on the bird images in CUB-200-2011~\cite{CUB}.

\paragraph{Estimating GT cue locations.}
We generate the ground-truth cue locations using following intuition: for two classes A and B differing only in one attribute $a$, the location $z$ for the cue for predicting A instead of B will correspond to the object part containing the attribute $a$. We explain algorithmically how we build the ground-truth attribution mask for an image $x$ with respect to two bird classes A and B in CUB. We first use the attribute annotations for 312 attributes in CUB to compute the set of attributes for each class: $\mathcal{S}^\text{A}$ and $\mathcal{S}^\text{B}$. For example, $\mathcal{S}^\text{Fish crow}=\{\text{black crown}, \text{black wing}, \text{all-purpose bill-shape},\cdots\}$. We then compute the symmetric difference of the attributes for the two classes $\mathcal{S}^\text{A}\triangle\mathcal{S}^\text{B}=(\mathcal{S}^\text{A}\cup\mathcal{S}^\text{B})\setminus(\mathcal{S}^\text{A}\cap\mathcal{S}^\text{B})$. Now, we map each attribute in $a\in\mathcal{S}^\text{A}\triangle\mathcal{S}^\text{B}$ to the corresponding bird part $p\in\mathcal{P}$ among 7 bird parts annotated in CUB. For example, the attribute-mismatching bird parts for classes ``Fish crow'' and ``Brandt cormorant'' are  $\mathcal{P}^{\text{A},\text{B}}=\{\text{head},\text{wing}\}$. We locate the parts $\mathcal{P}^{\text{A},\text{B}}$ in samples $x$ of classes A and B using the keypoint annotations in CUB: $\mathcal{K}^{\text{A},\text{B}}(x)$. We expand the keypoint annotations to a binary mask $\mathcal{M}^{\text{A},\text{B}}(x)\in\{0,1\}^{H\times W}$ using the nearest-neighbor assignment of pixels to bird parts. The final mask $\mathcal{M}^{\text{A},\text{B}}(x)$ for the input $x$ is used as the ground-truth attribution map. See the ``GT mask'' column in Figure~\ref{fig:interpretability-qualitative-cub} for example binary masks. For evaluation we use all class pairs in CUB with the number of attribute-differing parts $|\mathcal{P}^{\text{A},\text{B}}|\leq 3$, resulting in $31+64+96=191$ class pairs.

\paragraph{Counterfactual attributions.}
To predict the difference in needed cues for recognizing classes A and B, we obtain the {absolute values of} counterfactual attributions from each method by computing the difference $|s^\text{A}-s^\text{B}|\in[0,1]^{H\times W}$. The underlying assumption is that $s^\text{A}$ and $s^\text{B}$ point to cues corresponding to the attributes for A and B, respectively. Hence, by taking the difference, one removes the attributions on regions that are important for both A and B. 

\paragraph{Evaluation metric: mean pixel-wise AP.}
To measure how well attribution maps retrieve the ground-truth part pixels $\mathcal{M}^{\text{A},\text{B}}(x)$, we measure the average precision for the pixel retrieval task~\cite{PixelPrecisionRecall,WSOLEVALJOURNAL}. Given a threshold $\tau\in[0,1]$, we define the positive predictions as the set of pixels in over multiple images: $\{(n,h,w)\mid|s^\text{A}_{hw}(x_n)-s^\text{B}_{hw}(x_n)|\geq\tau\}$ for images $x_n$ from classes A and B. With the pixel-wise binary labels $\mathcal{M}_{hw}^{\text{A},\text{B}}(x_n)$, we compute the pixel-wise average precision (PxAP) for the class pair $(\text{A},\text{B})$ by computing the area under the precision-recall curve. We then take the mean of PxAP over all the class pairs of interest (\eg those with $|\mathcal{P}^{\text{A},\text{B}}|=1$) to compute the mPxAP.

\paragraph{Qualitative results.}
See Figure~\ref{fig:interpretability-qualitative-cub} for the qualitative examples of \ours and baselines including CAM. We observe that the counterfactual attribution maps $s^\text{A}-s^\text{B}$ generated by \oursem and \oursml are more accurate than CAM and gradient-based attribution methods; \oursem attributions are qualitatively more precise than \oursml. \ours tends to assign close-to-zero attributions on irrelevant regions, while the baseline methods tend to produce noisy attributions. The sparsity of \ours makes it qualitatively more interpretable than the baselines.

\paragraph{Quantitative results.} 
Table~\ref{tab:interpretability-part-retrieval} shows the mPxAP scores for \ours and baseline methods for retrieving relevant pixels as attribution regions. We examine CUB class pairs with the number of parts with attribute differences $|\mathcal{P}^{\text{A},\text{B}}|\in\{1,2,3\}$. We observe that \oursem outperforms the baselines in all three sets of class pairs, confirming the qualitative superiority of \oursem in Figure~\ref{fig:interpretability-qualitative-cub}. \oursem attains 4.7\%p better mPxAP than CAM on average over the three sets. \oursml tends to be sub-optimal, compared to \oursem (27.3\% vs 34.3\% mPxAP). The variants of gradients perform below a mere 20\% mPxAP on average. In conclusion, the counterfactual attribution by \ours generates precise localization of the important bird parts that matter for the recognition task.

\subsection{Remove-and-classify results}
\label{subsec:experiments-remove-and-classify}

One of the most widely used frameworks for evaluating the attribution task is the remove-and-classify evaluation~\cite{InterpretabilityEvalWojciech,PatternNet,RISE,LIME,InterpretabilityBenchmark}. Image pixels $x_{hw}$ are erased in descending order of importance dictated by the attribution values $s_{hw}$. We write $x_{- k}$ for the image where the top-$k$\% important pixels are removed. We use the meaningful perturbation of the ``blur'' type~\cite{VedaldiMeaingfulPerturbation} for erasing the pixels.
A good attribution method shall assign high attribution values on important pixels; erasing them will quickly drop the classification accuracy $\mathcal{A}_k$ with increasing $k$. We set the base reference accuracy $\mathcal{A}^r_k$ as the classifier's accuracy with $k$\% of the pixels erased at random. For each method, we report the relative accuracy $\mathcal{R}_k=\mathcal{A}_k/\mathcal{A}^r_k$ for different $k$.

\begin{figure}
    \centering
    \includegraphics[width=\linewidth]{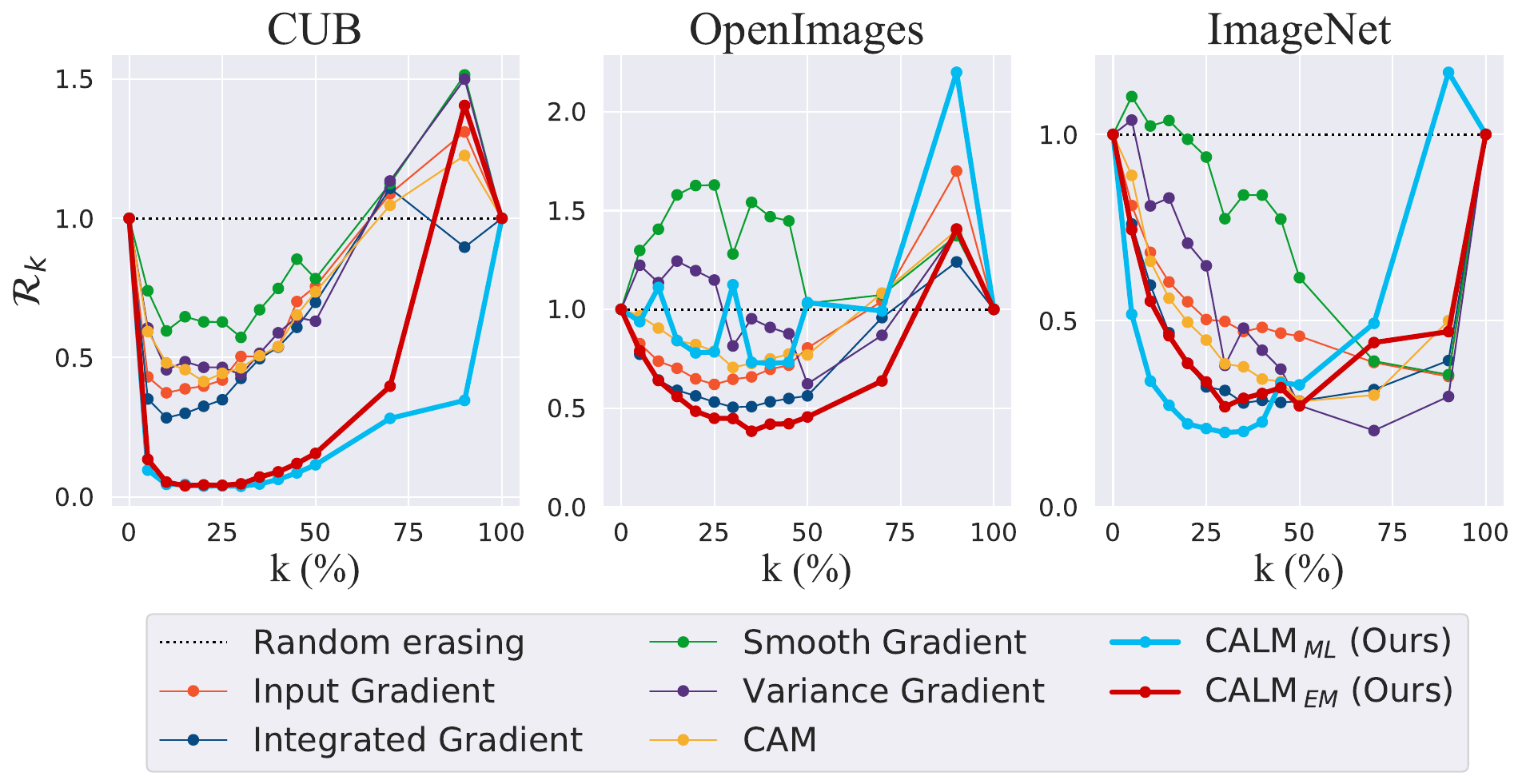}
    \caption{\small\textbf{Remove-and-classify results.} Classification accuracies of CNNs when $k$\% of pixels are erased according to the attribution values $s_{hw}$. We show the relative accuracies $\mathcal{R}_k$ against the random-erasing baseline. Lower is better.}
    \label{fig:interpretability-remove-and-classify}
    \vspace{-1em}
\end{figure}

\paragraph{Results.}
We show the remove-and-classify results in Figure~\ref{fig:interpretability-remove-and-classify} for three image classification datasets. We observe that \ours variants show the lowest relative accuracies (lower is better) $\mathcal{R}_k$ on cue-removed images in CUB and OpenImages, compared to CAM and other baselines. For the two datasets, \oursem attains values even close to zero at $k\in[10,50]$. On ImageNet, \oursem outperforms the baselines with a smaller margin. Overall, \oursem selects the important pixels for recognition best.

\setlength{\columnsep}{.3em}%
\begin{wraptable}{r}{.475\linewidth}
\setlength{\tabcolsep}{.15em}
\vspace{-1.2em}
    \centering
    \footnotesize
    \begin{tabular}{lcccc}
    Methods && CUB & Open & ImNet \\
	\cline{1-1}\cline{3-5}\vspace{-1em}\\
	\cline{1-1}\cline{3-5}\vspace{-1em} \\
	Baseline &  & 70.6 & 72.1 & 74.5  \\
    \oursem &  & 71.8 & 70.1 & 70.4  \\
    \oursml &  & 59.6 & 70.9 & 70.6  \\
    \cline{1-1}\cline{3-5}\vspace{-1em}\\
    \end{tabular}
    \caption{\footnotesize\textbf{Classification accuracy.}}
\vspace{-2em}

\end{wraptable}

\paragraph{Classification performances.}
We study the trade-off between interpretability and performance. The improved attribution performances come at the cost of decreased classification accuracies. 
Our models will be useful in applications that require great attribution performances at a small cost in model accuracies.

\subsection{Weakly-supervised object localization (WSOL)}
\label{subsec:experiments-wsol}

\begin{figure}
    \centering
    \includegraphics[width=.48\linewidth]{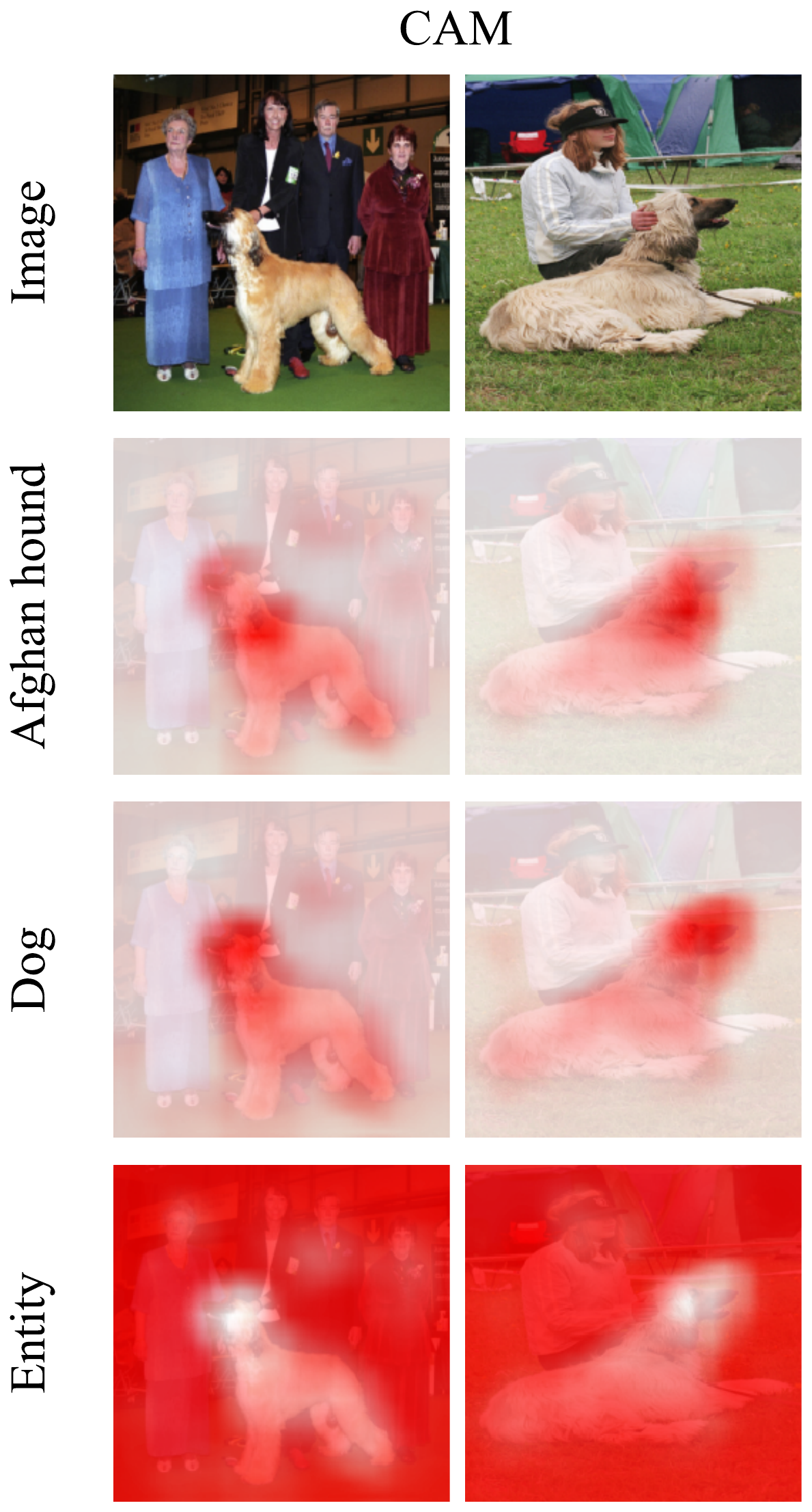}
    \includegraphics[width=.48\linewidth]{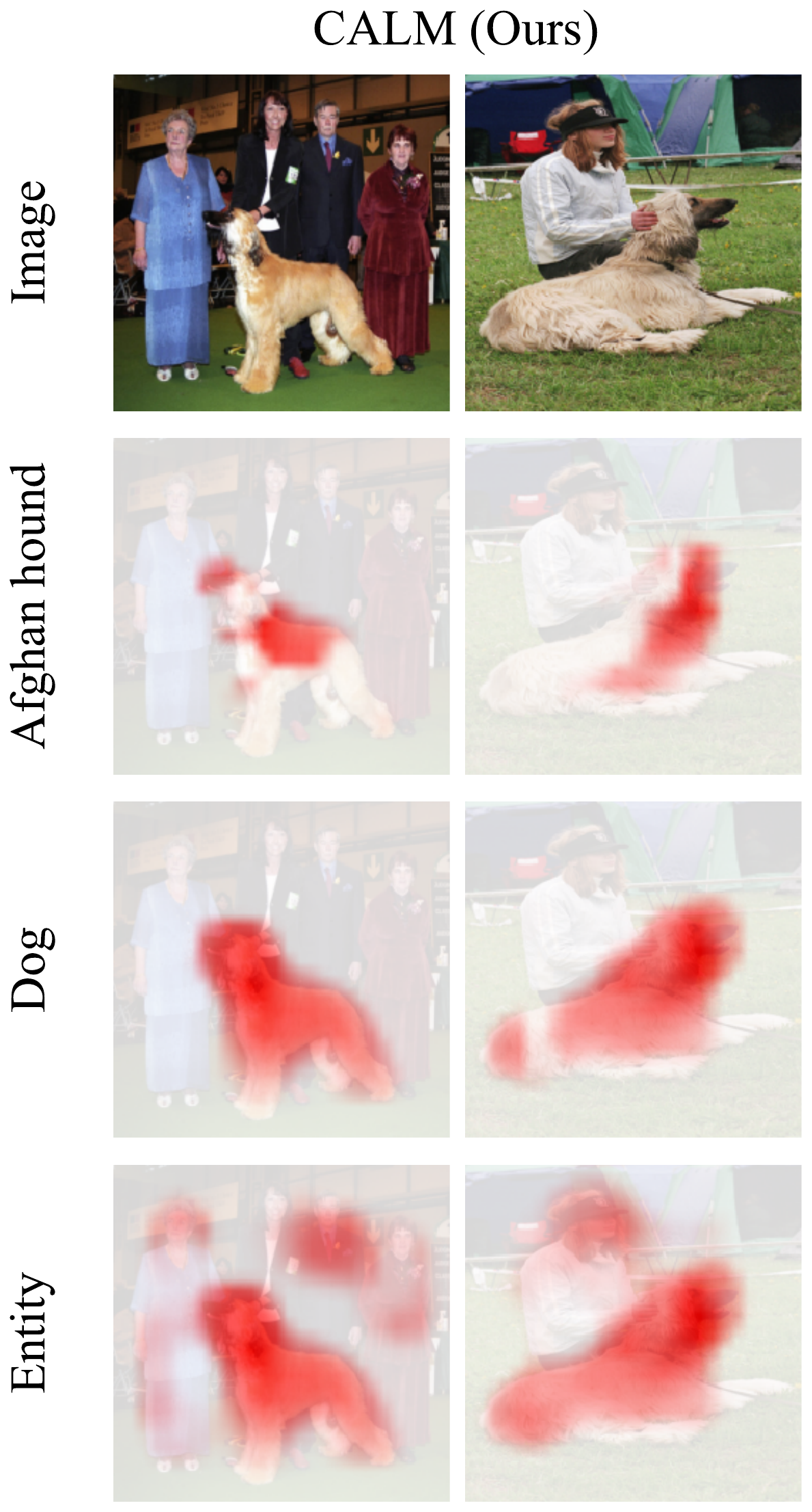}
    \caption{\small\textbf{Aggregating with class hierarchy.} ``Afghan hound'' is a descendant of ``Dog'', which is a descendant of ``Entity'' in the WordNet hierarchy. Selecting a sensible superset $\mathcal{Y}$ for aggregation lets \ours produce a high-quality foreground mask.}
    \label{fig:wsol-class-hierarchy}
    \vspace{-1em}
\end{figure}

\begin{table}[]
\setlength{\tabcolsep}{.5em}
    \centering
    \small
    \begin{tabular}{lcccc}
    Methods && ImageNet & CUB & OpenImages \\
	\cline{1-1}\cline{3-5}\vspace{-1em}\\
	\cline{1-1}\cline{3-5}\vspace{-1em} \\
	HaS~\cite{HaS} &  & 62.6 & 60.6 & 57.4 \\
	ACoL~\cite{ACoL} &  & 61.1 & 60.0 & 56.3 \\
	SPG~\cite{SPG} &  & 62.2 & 57.5 & 59.1 \\
	ADL~\cite{ADL} &  & 61.6 & 61.1 & 56.9 \\
	CutMix~\cite{CutMix} &  & 62.2 & 60.8 & 59.5 \\
	InCA~\cite{WSOL_INCA} &  & \textbf{63.1} & 63.4 & - \\
    \cline{1-1}\cline{3-5}\vspace{-1em}\\
	CAM~\cite{CAM} &  & 62.4 & 61.1 & 60.0  \\
	CAM~\cite{CAM} + $\mathcal{Y}$ &  & 60.6 & 63.4 & 60.0 \\
    \cline{1-1}\cline{3-5} \vspace{-1em}\\
    \oursem &  & 62.5 & 52.5 & \textbf{62.7}  \\
    \oursem + $\mathcal{Y}$ &  & 62.8 & 65.4 & \textbf{62.7} \\
    \oursml &  & 62.6 & 61.3 & 62.3  \\
    \oursml + $\mathcal{Y}$ &  & 62.7 & \textbf{68.0} & 62.3 \\
    \cline{1-1}\cline{3-5}\vspace{-1em}\\
    \end{tabular}
    
    \vspace{.5em}
    \caption{\small\textbf{WSOL results on CUB, OpenImages, and ImageNet.} Average for ResNet, Inception, and VGG are reported for each dataset. \oursem and \oursem+$\mathcal{Y}$ are compared against the baseline methods. ``+$\mathcal{Y}$'' denotes the aggregation.}
    \label{tab:wsol-results}
    \vspace{-1em}
\end{table}

WSOL is related to but different from the attribution task. For WSOL, one learns to detect \textit{object foreground regions} with only image-label pairs. While the ingredients are identical ($(X,Y)$ observed), the desired latent $Z$ is different: the important cues for recognition may not necessarily agree with the object foreground regions. Nonetheless, the WSOL field benefits from the developments in attribution methods like CAM, which has remained the state of the art method for WSOL for the past few years~\cite{WSOLEVALJOURNAL}.

We apply \ours to WSOL. Since attribution maps $p(\widehat{y},z|x)$ only point to sub-parts relevant for recognition, we aggregate the attributions them over multiple classes $p(\mathcal{Y},z|x)=\sum_{y\in\mathcal{Y}}p(y,z|x)$ (subset attribution in \S\ref{subsubsec:attribution-map-for-calm}) to fully cover the foreground regions. 

\paragraph{Setting the superset $\mathcal{Y}$.}
For the ground-truth class $\widehat{y}$, we set the superset $\mathcal{Y}$ as the set of classes sharing the same part composition as $\widehat{y}$. The intuition is that the attributions are mostly on object parts and that classes of such $\mathcal{Y}$ have attributions spread across different object parts. For example, all bird classes in CUB~\cite{CUB} shall share the same superset $\mathcal{Y}=\{\text{all 200 birds}\}$, as they share the same body part composition. On the other extreme, 100 classes in OpenImages~\cite{OpenImagesV5,WSOLEVALJOURNAL} do not share part structures across classes. Thus, we always set $\mathcal{Y}=\{\widehat{y}\}$. ImageNet1K~\cite{ImageNet} is mixed. Its 1000 classes include 116 dog species, but also many other objects and concepts that do not share the same part structure. For ImageNet, we have manually annotated the supersets $\mathcal{Y}$ for every class $\widehat{y}$, using the WordNet hierarchy~\cite{WordNet}. Details in the Supplementary Materials.

\paragraph{Results.}
We evaluate WSOL performances based on the benchmarks and evaluation metrics in \cite{WSOLEVALJOURNAL}. The benchmark considers 3 architectures (VGG, Inception, ResNet) and 3 datasets (CUB, OpenImages, ImageNet). Implementation details are in Supplementary Materials. We show results in Table~\ref{tab:wsol-results}. We observe that the aggregation significantly enhances the WSOL performances for \oursem: 52.5\% to 65.4\% on CUB. \oursem+$\mathcal{Y}$ attains the best performances on CUB and OpenImages and second-best on ImageNet.

\paragraph{Analysis.}
We study the 116 fine-grained dog species in ImageNet more closely. We show the aggregation of attribution maps in Figure~\ref{fig:wsol-class-hierarchy}. \ours for $\widehat{y}$ fails to cover the full extent of the object. As the maps are aggregated over all dog species $\mathcal{Y}$, the map precisely covers the full extents of the dogs. However, if $\mathcal{Y}$ covers all 1000 classes, the resulting saliency map $p(z|x)$ starts to include non-dog pixels.

\section{Conclusion}

Despite its great contributions to the field, the class activation mapping (CAM) is not as interpretable as it could be. It lacks communicability in practice and fails to meet key theoretical requirements for feature attribution methods. This paper has introduced a novel visual feature attribution method, class activation latent mapping (\ours). 
Based on the probabilistic treatment of the last layers of CNNs, \ours is interpretable by design. 
\ours satisfies the theoretical requirements as an attribution method and outperforms CAM and other baselines on attribution tasks.

\subsubsection*{Acknowledgement}
This work has been partially funded by the ERC (853489 - DEXIM) and by the DFG (2064/1 – Project number 390727645). We thank members of NAVER AI Lab, especially Sanghyuk Chun for their helpful feedback and discussion. Kay Choi designed Figure~\ref{fig:method-main}.

{\small
\bibliographystyle{ieee_fullname}
\bibliography{main}
}

\clearpage

\appendix

\section*{Supplementary Materials}

\noindent
Supplementary Materials contain supporting claims, derivations for formulae, and auxiliary experimental results for the materials in the main paper. The sections are composed sequentially with respect to the contents of the main paper.

\section{Equivalence of CAM formulations}

In \S\ref{sec:CAM} of the main paper, we have argued that the formulation in Equation~\ref{eq:cam-training} copied below,
\begin{align}
    p(y|x)= \text{softmax}\left(\frac{1}{HW}\sum_{hw}f_{yhw}(x)\right)
\end{align}
is equivalent to CNNs with an additional linear layer $W\in\mathbb{R}^{C\times L}$ after the global average pooling (e.g. ResNet):
\begin{align}
    p(y|x)= \text{softmax}\left(\sum_{l}W_{yl}\left(\frac{1}{HW}\sum_{hw}\overline{f}_{lhw}(x)\right)\right)
    \label{appendix:eq:original-cam}
\end{align}
where $\overline{f}$ is a fully-convolutional network with output dimensionality $\overline{f}(x)\in\mathbb{R}^{L\times H\times C}$.
This follows from the distributive property of sums and multiplications:
\begin{align}
    \sum_{l}W_{yl}\left(\frac{1}{HW}\sum_{hw}\overline{f}_{lhw}(x)\right)=\frac{1}{HW}\sum_{hw}\sum_{l}W_{yl}\overline{f}_{lhw}(x)
\end{align}
where we may re-define $f_{yhw}:= \sum_{l}W_{yl}\overline{f}_{lhw}$ as another fully-convolutional network with a convolutional layer with $1\times 1$ kernels ($W_{yl}$) at the end.

For networks of the form in Equation~\ref{appendix:eq:original-cam}, the original CAM algorithm computes the attribution map by first taking the sum
\begin{align}
    {f}_{hw}=\sum_l W_{\widehat{y}l}\overline{f}_{lhw}(x)
\end{align}
and normalizing ${f}$ as in Equation~\ref{eq:scoremap-maxnorm} in main paper:
\begin{align}
    s=
    \begin{cases}
        ({f}^{\widehat{y}}_{\max})^{-1}\max(0,{f}^{\widehat{y}}) & \text{max~\cite{CAM}} \\
        ({f}^{\widehat{y}}_{\max}-{f}^{\widehat{y}}_{\min})^{-1}({f}^{\widehat{y}}-{f}^{\widehat{y}}_{\min}) & \text{min-max~\cite{GRADCAM} }
    \end{cases}
\label{appendix:eq:scoremap-maxnorm}
\end{align}
Hence, for both training and interpretation, the family of architectures described by Equation~\ref{eq:cam-training} subsumes the family originally considered in CAM~\cite{CAM} (Equation~\ref{appendix:eq:original-cam}).


\section{Derivation of \oursem objective}

In \S\ref{sec:training_algorithms}, we have introduced an expectation-maximization (EM) learning framework for our latent variable model. We derive the EM objective in Equation~\ref{eq:calm-em-implementation} here. Our aim is to minimize the negative log-likelihood $-\log p_\theta ({y}|{x})$. We upper bound the objective as follows.
\begin{align}
    -\log p_\theta({y}|{x})&=-\log \sum_z p_\theta({y},z|{x}) \\
    &=-\log \sum_z p_{\theta^\prime}(z|{x},{y}) \frac{p_\theta({y},z|{x})}{p_{\theta^\prime}(z|{x},{y})} \label{pre-jensen}\\
    &\leq-\sum_z p_{\theta^\prime}(z|{x},{y})\log \frac{p_\theta({y},z|x)}{p_{\theta^\prime}(z|{x},{y})} \label{post-jensen}\\
    &\leq-\sum_z p_{\theta^\prime}(z|{x},{y})\log p_\theta({y},z|{x}) \, . \label{appendix:eq:derivation-EM}
\end{align}
{The inequalities leading to Equation~\ref{post-jensen} and \ref{appendix:eq:derivation-EM} follow from the Jensen's inequality} and the positivity of the entropy, respectively.

We parametrize each term with a neural network. $p_\theta({y},z|{x})$ is computed via ${g}_{yz}\cdot{h}_z$ and $p_{\theta^\prime}(z|{x},{y})$ is first decomposed as
\begin{align}
    p_{\theta^\prime}(z|{x},{y})=\frac{p_{\theta^\prime}({y},z|{x})}{\sum_{l} p_{\theta^\prime}({y},l|{x})}
\end{align}
and computed with neural networks
\begin{align}
    p_{\theta^\prime}(z|{x},{y})=\frac{{g}^\prime_{yz}\cdot{h}^\prime_z}{\sum_{l} {g}^\prime_{yl}\cdot{h}^\prime_l}
\end{align}
where $g^\prime$ and $h^\prime$ are neural networks parametrized with $\theta^\prime$.



\section{Training details for \ours}

We provide miscellaneous training details for \ours. See \S\ref{subsec:baseline-attribution-methods} in the main paper for major training details.

\paragraph{Architecture.} We use the ResNet50 as the feature extractor. We enlarge the attribution map size to $28 \times 28$ by changing the stride of the last two residual blocks from $2$ to $1$. Two CNN branches $g$ and $h$ are followed by the feature extractor. The branch $g$ is the one convolutional layer of kernel size $1$ and stride $1$ with the number of output channel to be the number of classes. The branch $h$ is composed of one convolutional layer of kernel size $1$ and stride $1$ with the number of output channel to be $1$, followed by the ReLU activation function.

\paragraph{Optimization hyperparameters.}
We use the stochastic gradient descent with the momentum $0.9$ and weight decay $1\times10^{-4}$. We set the learning rate as ($3\times10^{-5}$, $5\times10^{-5}$, $7\times10^{-5}$) for (CUB, OpenImages, ImageNet).

\section{More qualitative results}

\begin{figure*}
    \centering
    \includegraphics[width=\linewidth]{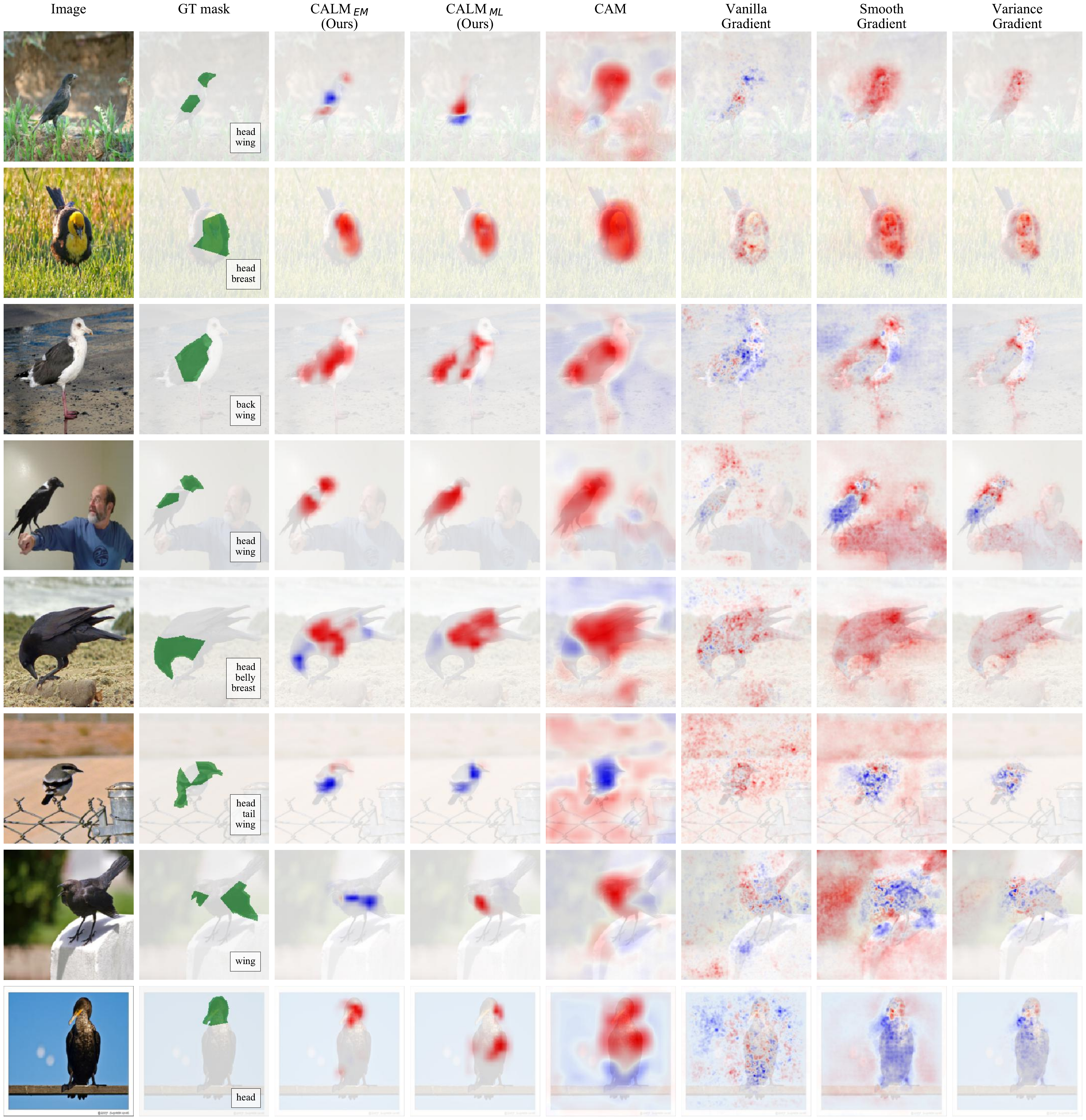}
    \caption{\small\textbf{Counterfactual attribution maps on CUB.} Extension of Figure~\textcolor{red}{4} in main paper. We compare the counterfactual attributions from CALM and baseline methods against the GT attribution mask. The GT mask indicates the bird parts where the attributes for the class pair (A,B) differ. The counterfactual attributions denote the difference between the maps for classes A and B: $s^\text{A}-s^\text{B}$. Red: positive values. Blue: negative values.}
    \label{appendix:fig:interpretability-qualitative-counterfactual-cub}
\end{figure*}

See qualitative results in Figure~\ref{appendix:fig:interpretability-qualitative-counterfactual-cub} for the comparison of attributions against the ground-truth attributions using the counterfactual maps $s^\text{A}-s^\text{B}$. As in Figure~\ref{fig:interpretability-qualitative-cub} of the main paper, we observe that \ours attains the best similarity with the ground-truth masks. \ours are also the most human-understandable among the considered methods.

\begin{figure*}
    \centering
    \includegraphics[width=\linewidth]{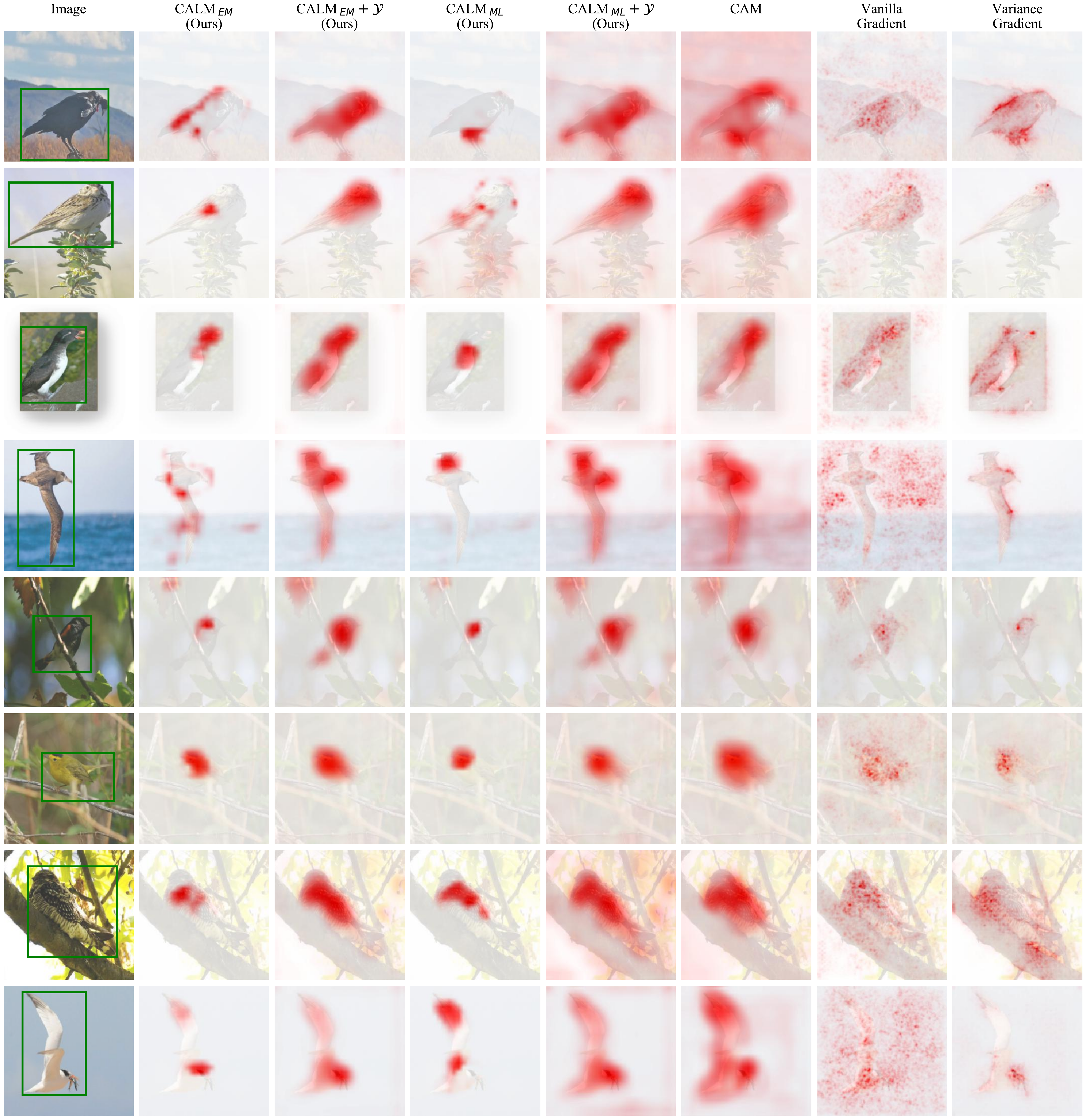}
    \caption{\small\textbf{Examples of attribution maps on CUB.} We show the object bounding boxes to mark the foreground regions. +$\mathcal{Y}$ denotes the class aggregation.}
    \label{appendix:fig:interpretability-qualitative-cub}
\end{figure*}

\begin{figure*}
    \centering
    \includegraphics[width=\linewidth]{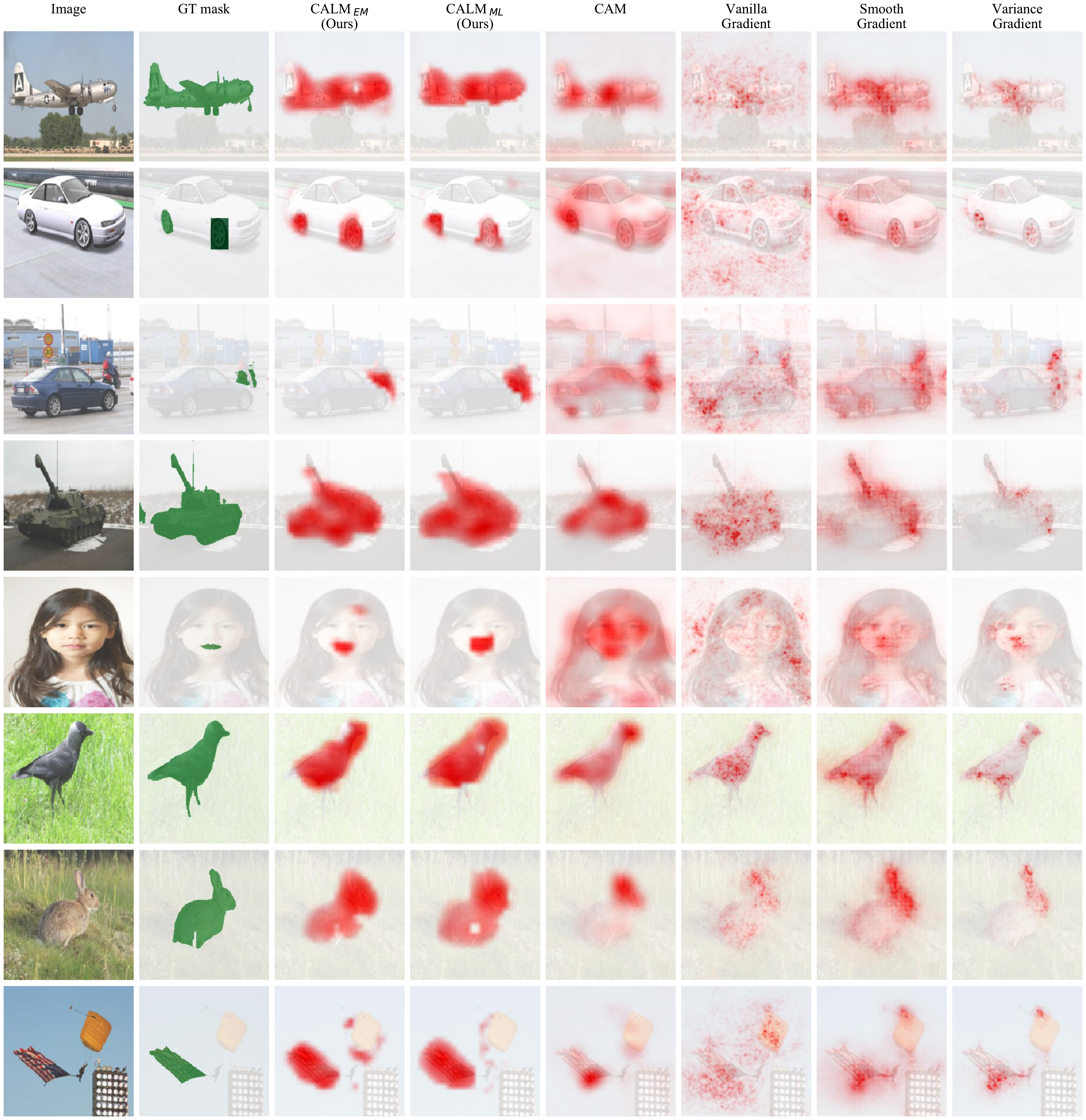}
    \caption{\small\textbf{Examples of attribution maps on OpenImages.} We show the object bounding boxes to mark the foreground regions. We do not show the class aggregation (+$\mathcal{Y}$) because it does not change our methods (\oursem and \oursml) on OpenImages ($\mathcal{Y}$ are all singletons).}
    \label{appendix:fig:interpretability-qualitative-openimages}
\end{figure*}

\begin{figure*}
    \centering
    \includegraphics[width=\linewidth]{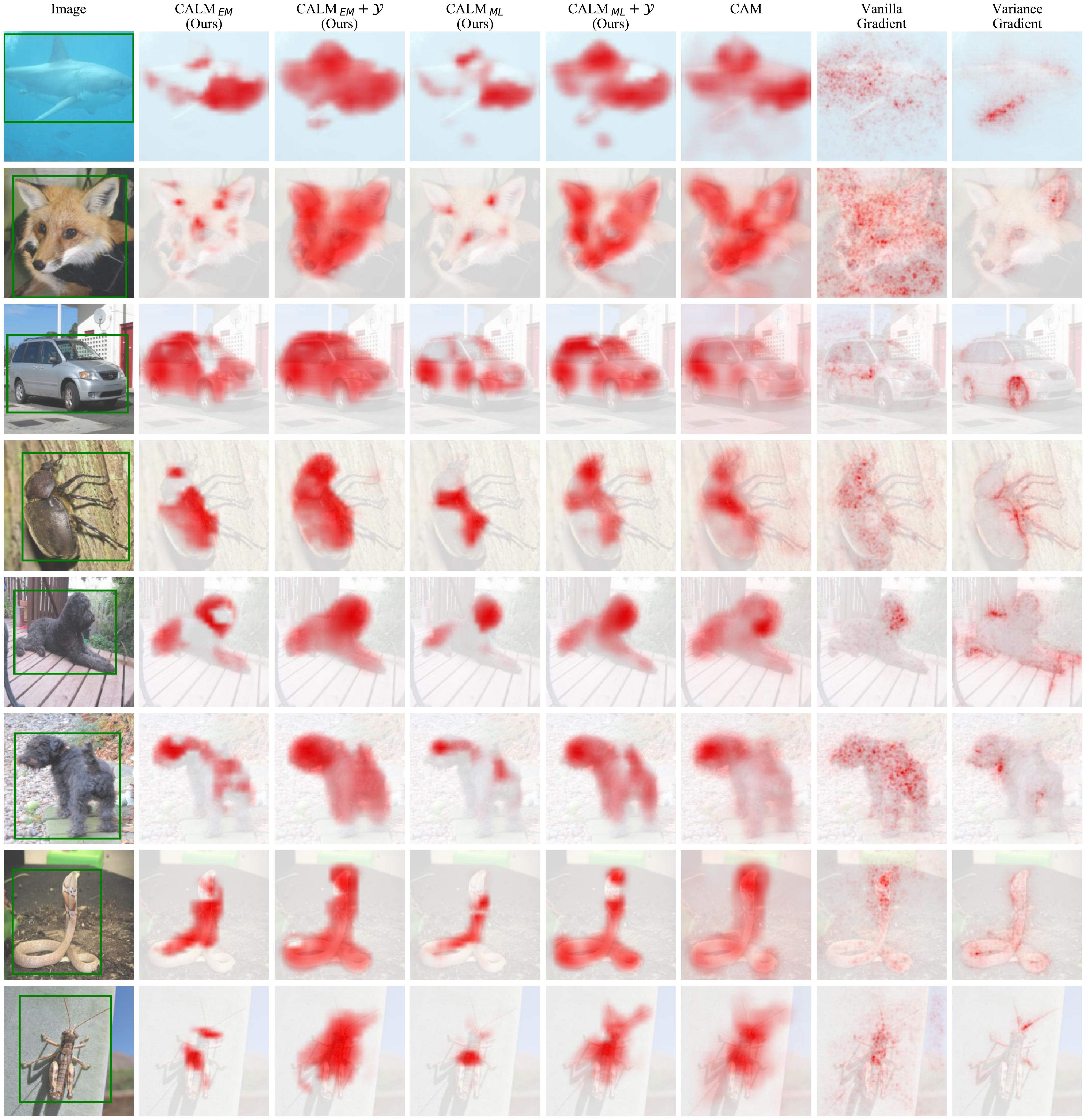}
    \caption{\small\textbf{Examples of attribution maps on ImageNet.} We show the object bounding boxes to mark the foreground regions. +$\mathcal{Y}$ denotes the class aggregation.}
    \label{appendix:fig:interpretability-qualitative-imagenet}
\end{figure*}

We add more qualitative results of the attribution maps $s^\text{A}$ for the ground-truth class for CUB (Figure~\ref{appendix:fig:interpretability-qualitative-cub}), OpenImages (Figure~\ref{appendix:fig:interpretability-qualitative-openimages}), and ImageNet (Figure~\ref{appendix:fig:interpretability-qualitative-imagenet}). For each case, we do not have the \textit{GT attribution maps} as in Figure~\ref{appendix:fig:interpretability-qualitative-counterfactual-cub}, but we show the \textit{GT foreground} object bounding boxes, or masks if available (OpenImages). Note that the attribution maps (\oursem and \oursml) are not designed to highlight the full object extent, while the aggregated versions (+$\mathcal{Y}$) are designed so. We observe that in all three datasets, \oursem+$\mathcal{Y}$ and \oursml+$\mathcal{Y}$ tend to generate high-quality foreground masks for the object of interest; for OpenImages, note that ``+$\mathcal{Y}$'' does not change \oursem and \oursml as the supersets $\mathcal{Y}$ are all singletons.

\section{Evaluation protocol for WSOL}

In \S\ref{subsec:experiments-wsol} of the main paper, we have presented experimental results on WSOL benchmarks. In this section, we present metrics, datasets, and validation protocols for the WSOL experiments. We follow the recently proposed WSOL evaluation protocol~\cite{WSOLEVAL}.

\noindent \textbf{Metrics.} After computing the attribution $s$, WSOL methods binarize the attribution by a threshold $\tau$ to generate a foreground mask $\mathbf{M}=\mathbbm{1}[s_{ij}\geq \tau]$. When there are mask annotations in the dataset, we compute pixel-wise precision and recall at the threshold $\tau$. The \pxap is the area under the precision and recall curve at all possible thresholds $\tau\in[0,1]$. On the other hand, when only the box annotations are available, we generate a bounding box that tightly bound each connected component on the mask $\mathbf{M}$. Then, we calculate intersection over union (IoU) between all pairs of ground truth boxes and predicted boxes at all thresholds $\tau\in[0,1]$. When there is at least one pair of (ground truth, prediction, $\tau$) with $\text{IoU}\geq \delta$, we regard the localization prediction of the attribution as correct. The \maxboxacc is the average of three ratios of correct images in the dataset at three $\delta=0.3, 0.5, 0.7$.

\noindent \textbf{Evaluation protocol.} Every dataset in our WSOL experiments consists of three disjoint splits: \trainweaksup, \trainfullsup, and \testfullsup. The \trainfullsup and \testfullsup splits contain images with localization and class labels, while the \trainweaksup split contains images only with class labels. We use the \trainweaksup split to train the classifier, and tune our hyperparameter by checking the localization performance on the \trainfullsup split. Specifically, we only tune the learning rate by randomly sampling 30 learning rates from the log-uniform distribution $\text{LogUniform}[5^{-5},5^{-1}]$. Then, based on the performance on \trainfullsup, we select the optimal learning rate. Finally, we measure the final localization performance on \testfullsup split with the selected learning rate. Since previous WSOL methods in~\cite{WSOLEVAL} are also evaluated under this evaluation protocol, we can compare fairly our method with the previous methods.

\section{Superset $\mathcal{Y}$ for ImageNet classes}

In the main paper \S\ref{subsec:experiments-wsol}, we have discussed the disjoint goals pursued by two tasks, visual attribution and WSOL. The former aims to locate cues that make the class distinguished from the others and typically locates a small part of the object; the latter aims to locate the foreground pixels for objects. To bridge the two, we have considered an aggregation strategy to produce a foreground mask from attribution maps. Our assumption is that attribution maps for different classes for the same family sharing the identical object structure will cover various object parts in the foreground mask. For example, 200 bird classes in CUB share the part structure of ``head-beak-breast-wing-belly-leg-tail'', but each class will highlight different parts. Combining the attribution maps for the 200 classes will roughly cover all the object parts, providing a higher-quality foreground mask. 

For 1000 classes in ImageNet1K, we manually annotate the corresponding supersets $\mathcal{Y}$. We first build a tree of concepts over the classes using the WordNet hierarchy~\cite{WordNet}. The leaf nodes correspond to 1000 classes, while the root node corresponds to the concept ``entity'' that includes all 1000 classes. For each leaf node (ImageNet class) $y$, we have annotated the superset $\mathcal{Y}$ by choosing an appropriate parent node of $y$. The parent selection criteria is as follows:
\begin{itemize}
    \item Choose the parent node $\text{PA}(y)$ of $y$ as close to the root as possible;
    \item Such that $\mathcal{Y}$, the set of all children of $\text{PA}(y)$, consists of classes with the same object structure as $y$.
\end{itemize}

\paragraph{Algorithm.}
We efficiently annotate the parent nodes by traversing the tree in a breadth-first-search (BFS) manner from the root node, ``entity''. We start from the direct children (depth$=1$) of the root node. For each node of depth 1, we mark if the concept contains classes of the same object structure (\texttt{hom} or \texttt{het}). For example, the family of classes under the \texttt{organism} parent is not yet specific enough to contain classes of homogeneous object structures, so we mark \texttt{het}. We continue traversing in depths 2, 3, and so on. If a node at depth $d$ is marked \texttt{hom}, we treat all of its children to have the same superset $\mathcal{Y}$ and do not traverse its descendants for depth$>d$. For example, the \texttt{canine} superset of 116 ImageNet classes is reached by following the genealogy of \texttt{organism} \textrightarrow~\texttt{animal} \textrightarrow~\texttt{chordate} \textrightarrow~\texttt{mammal} \textrightarrow~\texttt{placental} \textrightarrow~\texttt{carnivore} \textrightarrow~\texttt{canine}. 
We note that the task is fairly well-defined for humans.

\paragraph{Results.}
We find 450 supersets in ImageNet1K. 120 of them are non-singleton, consisting of at least two ImageNet classes. The rest 330 classes are singleton supersets. See Table~\ref{tab:wsol-superset} for the list. 

\section{A complete version of WSOL results}
See Table~\ref{appendix:tab:wsol-results} for the complete version of Table~\ref{tab:wsol-results} of the main paper. We show the architecture-wise performances for \ours, CAM, and six previous WSOL methods (HaS~\cite{HaS}, ACoL~\cite{ACoL}, SPG~\cite{SPG}, ADL~\cite{ADL}, CutMix~\cite{CutMix}, InCA~\cite{WSOL_INCA}).

In the ImageNet1K dataset, the proposed method achieves competitive performance (62.8\%) with CAM (62.4\%) and InCA (63.1\%). The superset $\mathcal{Y}$ aggregation improves the localization performances for \oursem (62.5\% \textrightarrow~ 62.8\%), but it decreases the CAM performances significantly (62.4\% \textrightarrow~ 60.6\%). For CUB, we observe that \oursem does not localize the birds effectively without the superset $\mathcal{Y}$ (52.5\%). With the superset aggregation, \oursem attains 65.4\%. This is expected behavior because \oursem attributions often highlight small discriminative object parts (\eg Figure~\ref{fig:interpretability-qualitative-cub} in the main paper). For OpenImages, \oursem achieves the state-of-the-art performances on all three backbone architectures (61.3\%, 64.4\%, 62.5\%). Since the modified OpenImages~\cite{WSOLEVAL} consists of classes of unique object structures, the supersets are all singletons. The performances are thus identical with or without the aggregation $+\mathcal{Y}$.
In summary, \oursem on ImageNet is competitive, compared to the state of the art, and is the new state of the art on CUB and OpenImages.

\section{More qualitative examples for WSOL}

\begin{figure*}
    \centering
    \includegraphics[width=.33\linewidth]{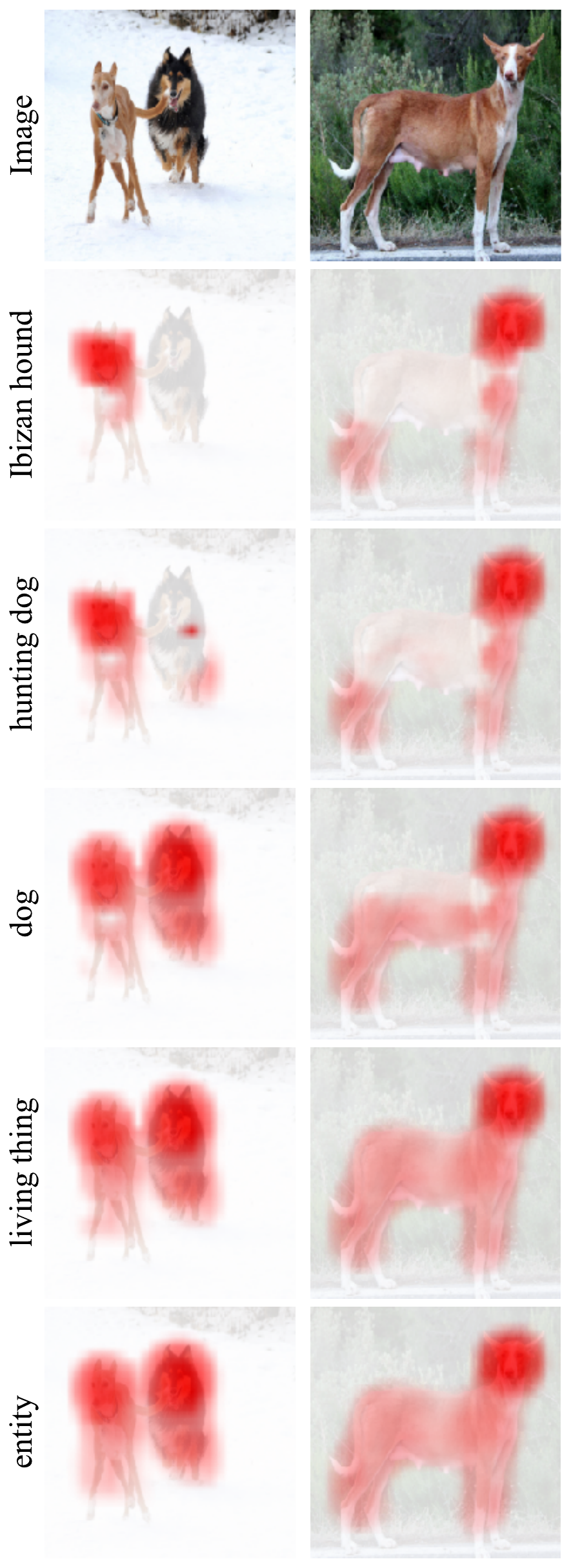}
    \includegraphics[width=.33\linewidth]{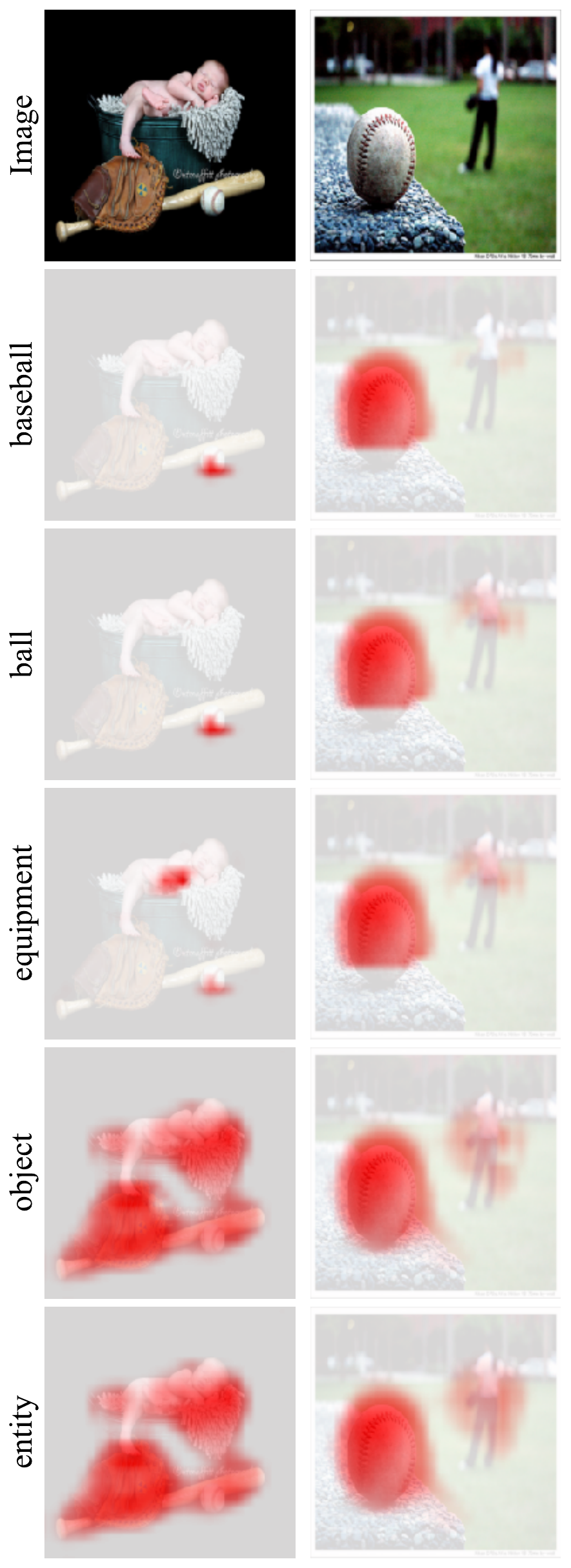}
    \includegraphics[width=.33\linewidth]{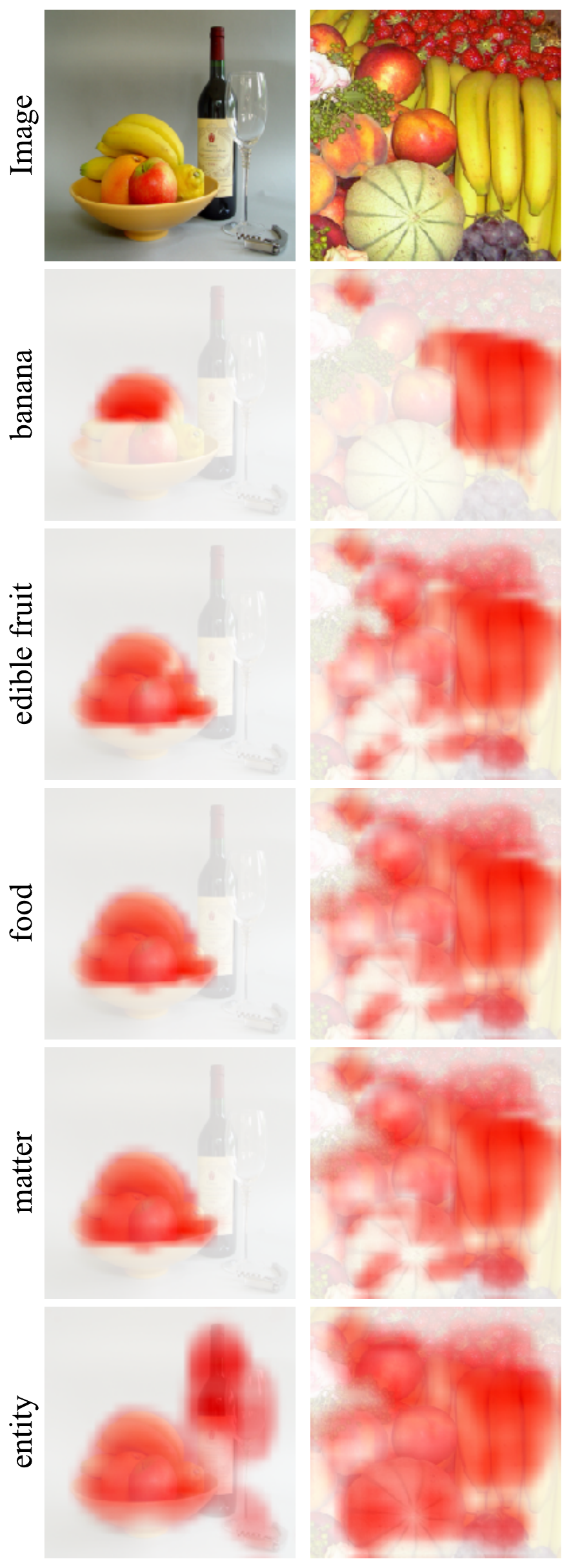}
    \caption{\small\textbf{Examples for aggregation at different hierarchies on ImageNet.} Extension of Figure~\textcolor{red}{6} of main paper. We show the aggregated attribution maps at different depths of the hierarchy and the correspondingly expanding superset $\mathcal{Y}$.}
    \label{appendix:fig:more-wsol-superset-qualitative}
\end{figure*}

In Figure~\ref{fig:wsol-class-hierarchy} of the main paper, we have shown the aggregation of attribution maps at different depths of hierarchy for the ``canine'' class. We show additional qualitative examples of the superset aggregation for other classes in Figure~\ref{appendix:fig:more-wsol-superset-qualitative}. We note that the optimal depths that precisely cover the object extents differ across classes.

\begin{table*}[]
\setlength{\tabcolsep}{.2em}
\centering
\footnotesize
\begin{tabular}{lccclccclcc}
Class name & WordNet ID & \# Classes & & Class name & WordNet ID & \# Classes & & Class name & WordNet ID & \# Classes \\
\cline{1-3}\cline{5-7}\cline{9-11}\vspace{-1em} \\
\cline{1-3}\cline{5-7}\cline{9-11}\vspace{-1em} \\
canine		&	\texttt{	n02083346	}	&	116	&	&	wheel	&	\texttt{	n04574999	}	&	4	&	&	power tool	&	\texttt{	n03997484	}	&	2	\\
bird	&	\texttt{	n01503061	}	&	52	&	&	swine	&	\texttt{	n02395003	}	&	3	&	&	farm machine	&	\texttt{	n03322940	}	&	2	\\
reptile	&	\texttt{	n01661091	}	&	36	&	&	lagomorph	&	\texttt{	n02323449	}	&	3	&	&	slot machine	&	\texttt{	n04243941	}	&	2	\\
insect	&	\texttt{	n02159955	}	&	27	&	&	marsupial	&	\texttt{	n01874434	}	&	3	&	&	free-reed instrument	&	\texttt{	n03393324	}	&	2	\\
primate	&	\texttt{	n02469914	}	&	20	&	&	coelenterate	&	\texttt{	n01909422	}	&	3	&	&	piano	&	\texttt{	n03928116	}	&	2	\\
ungulate	&	\texttt{	n02370806	}	&	17	&	&	echinoderm	&	\texttt{	n02316707	}	&	3	&	&	lock	&	\texttt{	n03682487	}	&	2	\\
aquatic vertebrate	&	\texttt{	n01473806	}	&	16	&	&	person	&	\texttt{	n00007846	}	&	3	&	&	breathing device	&	\texttt{	n02895606	}	&	2	\\
building	&	\texttt{	n02913152	}	&	12	&	&	firearm	&	\texttt{	n03343853	}	&	3	&	&	heater	&	\texttt{	n03508101	}	&	2	\\
car	&	\texttt{	n02958343	}	&	10	&	&	clock	&	\texttt{	n03046257	}	&	3	&	&	locomotive	&	\texttt{	n03684823	}	&	2	\\
bovid	&	\texttt{	n02401031	}	&	9	&	&	portable computer	&	\texttt{	n03985232	}	&	3	&	&	bicycle,	&	\texttt{	n02834778	}	&	2	\\
arachnid	&	\texttt{	n01769347	}	&	9	&	&	brass	&	\texttt{	n02891788	}	&	3	&	&	railcar	&	\texttt{	n02959942	}	&	2	\\
ball	&	\texttt{	n02778669	}	&	9	&	&	cart	&	\texttt{	n02970849	}	&	3	&	&	handcart	&	\texttt{	n03484083	}	&	2	\\
headdress	&	\texttt{	n03502509	}	&	9	&	&	sailing vessel	&	\texttt{	n04128837	}	&	3	&	&	warship	&	\texttt{	n04552696	}	&	2	\\
feline	&	\texttt{	n02120997	}	&	8	&	&	aircraft	&	\texttt{	n02686568	}	&	3	&	&	sled	&	\texttt{	n04235291	}	&	2	\\
amphibian	&	\texttt{	n01627424	}	&	8	&	&	bus	&	\texttt{	n02924116	}	&	3	&	&	reservoir	&	\texttt{	n04078574	}	&	2	\\
decapod crustacean	&	\texttt{	n01976146	}	&	8	&	&	pot	&	\texttt{	n03990474	}	&	3	&	&	jar	&	\texttt{	n03593526	}	&	2	\\
place of business	&	\texttt{	n03953020	}	&	8	&	&	dish	&	\texttt{	n03206908	}	&	3	&	&	basket	&	\texttt{	n02801938	}	&	2	\\
musteline mammal	&	\texttt{	n02441326	}	&	7	&	&	pot	&	\texttt{	n03990474	}	&	3	&	&	glass	&	\texttt{	n03438257	}	&	2	\\
fungus	&	\texttt{	n12992868	}	&	7	&	&	pen	&	\texttt{	n03906997	}	&	3	&	&	shaker	&	\texttt{	n04183329	}	&	2	\\
truck	&	\texttt{	n04490091	}	&	7	&	&	telephone, phone	&	\texttt{	n04401088	}	&	3	&	&	opener	&	\texttt{	n03848348	}	&	2	\\
bottle	&	\texttt{	n02876657	}	&	7	&	&	gymnastic apparatus	&	\texttt{	n03472232	}	&	3	&	&	power tool	&	\texttt{	n03997484	}	&	2	\\
seat	&	\texttt{	n04161981	}	&	7	&	&	neckwear	&	\texttt{	n03816005	}	&	3	&	&	pan, cooking pan	&	\texttt{	n03880531	}	&	2	\\
rodent	&	\texttt{	n02329401	}	&	6	&	&	swimsuit	&	\texttt{	n04371563	}	&	3	&	&	cleaning implement	&	\texttt{	n03039947	}	&	2	\\
mollusk	&	\texttt{	n01940736	}	&	6	&	&	body armor	&	\texttt{	n02862048	}	&	3	&	&	puzzle	&	\texttt{	n04028315	}	&	2	\\
stringed instrument	&	\texttt{	n04338517	}	&	6	&	&	footwear	&	\texttt{	n03381126	}	&	3	&	&	camera	&	\texttt{	n02942699	}	&	2	\\
boat	&	\texttt{	n02858304	}	&	6	&	&	bridge	&	\texttt{	n02898711	}	&	3	&	&	weight	&	\texttt{	n04571292	}	&	2	\\
box	&	\texttt{	n02883344	}	&	6	&	&	memorial	&	\texttt{	n03743902	}	&	3	&	&	cabinet	&	\texttt{	n02933112	}	&	2	\\
toiletry	&	\texttt{	n04447443	}	&	6	&	&	alcohol	&	\texttt{	n07884567	}	&	3	&	&	curtain	&	\texttt{	n03151077	}	&	2	\\
bag	&	\texttt{	n02773037	}	&	5	&	&	dessert	&	\texttt{	n07609840	}	&	3	&	&	sweater	&	\texttt{	n04370048	}	&	2	\\
stick	&	\texttt{	n04317420	}	&	5	&	&	cruciferous vegetable	&	\texttt{	n07713395	}	&	3	&	&	robe	&	\texttt{	n04097866	}	&	2	\\
bear	&	\texttt{	n02131653	}	&	4	&	&	baby bed	&	\texttt{	n02766320	}	&	3	&	&	scarf	&	\texttt{	n04143897	}	&	2	\\
woodwind	&	\texttt{	n04598582	}	&	4	&	&	procyonid	&	\texttt{	n02507649	}	&	2	&	&	gown	&	\texttt{	n03450516	}	&	2	\\
source of illumination	&	\texttt{	n04263760	}	&	4	&	&	viverrine	&	\texttt{	n02134971	}	&	2	&	&	protective garment	&	\texttt{	n04015204	}	&	2	\\
ship	&	\texttt{	n04194289	}	&	4	&	&	cetacean	&	\texttt{	n02062430	}	&	2	&	&	oven	&	\texttt{	n03862676	}	&	2	\\
edge tool	&	\texttt{	n03265032	}	&	4	&	&	edentate	&	\texttt{	n02453611	}	&	2	&	&	sheath	&	\texttt{	n04187061	}	&	2	\\
overgarment	&	\texttt{	n03863923	}	&	4	&	&	elephant	&	\texttt{	n02503517	}	&	2	&	&	movable barrier	&	\texttt{	n03795580	}	&	2	\\
skirt	&	\texttt{	n04230808	}	&	4	&	&	prototherian	&	\texttt{	n01871543	}	&	2	&	&	sheet	&	\texttt{	n04188643	}	&	2	\\
roof	&	\texttt{	n04105068	}	&	4	&	&	worm	&	\texttt{	n01922303	}	&	2	&	&	plaything	&	\texttt{	n03964744	}	&	2	\\
fence	&	\texttt{	n03327234	}	&	4	&	&	flower	&	\texttt{	n11669921	}	&	2	&	&	mountain	&	\texttt{	n09359803	}	&	2	\\
piece of cloth	&	\texttt{	n03932670	}	&	4	&	&	timer	&	\texttt{	n04438304	}	&	2	&	&	shore	&	\texttt{	n09433442	}	&	2	\\
\cline{1-3}\cline{5-7}\cline{9-11}\vspace{-1em}
\end{tabular}
\vspace{.5em}
\caption{\small\textbf{List of supersets $\mathcal{Y_{\text{fine}}}$.} We list 120 supersets for classes in ImageNet1K. We omit the rest 330 supersets from the list, as they only have single elements (singletons).}
\label{tab:wsol-superset}
\end{table*}

\definecolor{Gray}{gray}{0.85}
\newcolumntype{g}{>{\columncolor{Gray}}c}

\begin{table*}[]
\setlength{\tabcolsep}{.3em}
    \centering
    \small
    \begin{tabular}{lc*{3}{c}gc*{3}{c}gc*{3}{c}gcg}
    & & \multicolumn{4}{c}{ImageNet (\maxboxacc)} & & \multicolumn{4}{c}{CUB (\maxboxacc)}  & & \multicolumn{4}{c}{OpenImages (\pxap)} & & Total \\
	Methods  &  & VGG & Inception & ResNet & Mean &  & VGG & Inception & ResNet & Mean &  & VGG & Inception & ResNet & Mean & & Mean\\
	\cline{1-1}\cline{3-6}\cline{8-11}\cline{13-16}\cline{18-18} \vspace{-1em} \\
	\cline{1-1}\cline{3-6}\cline{8-11}\cline{13-16}\cline{18-18} \vspace{-1em} \\
	HaS~\cite{HaS} &  & 60.6 & \textbf{63.7} & 63.4 & {62.6} &  & 63.7 & 53.4 & 64.6 & {60.6} &  & 58.1 & 58.1 & 55.9 & {57.4} && {60.2}\\
	ACoL~\cite{ACoL} &  & 57.4 & \textbf{63.7} & 62.3 & {61.1} &  & 57.4 & 56.2 & 66.4 & {60.0} &  & 54.3 & 57.2 & 57.3 & {56.3} && {59.1} \\
	SPG~\cite{SPG} &  & 59.9 & 63.3 & 63.3 & {62.2} &  & 56.3 & 55.9 & 60.4 & {57.5} &  & 58.3 & 62.3 & 56.7 & {59.1} && {59.6} \\
	ADL~\cite{ADL} &  & 59.9 & 61.4 & 63.7 & {61.6} &  & 66.3 & 58.8 & 58.3 & {61.1} &  & 58.7 & 56.9 & 55.2 & {56.9} && {59.9} \\
	CutMix~\cite{CutMix} &  & 59.5 & 63.9 & 63.3 & {62.2} &  & 62.3 & 57.4 & 62.8 & {60.8} &  & 58.1 & 62.6 & 57.7 & {59.5} && {60.8}\\
	InCA~\cite{WSOL_INCA} &  & 61.3 & 62.8 & \textbf{65.1} & \textbf{63.1} &  & \textbf{66.7} & \textbf{60.3} & 63.2 & {63.4} &  & - & - & - & - && {-}\\
    \cline{1-1}\cline{3-6}\cline{8-11}\cline{13-16}\cline{18-18} \vspace{-1em}\\
	CAM~\cite{CAM} &  & 60.0 & 63.4 & 63.7 & {62.4} &  & 63.7 & 56.7 & 63.0 & {61.1} &  & 58.3 & 63.2 & 58.5 & {60.0} && {61.2} \\
	CAM~\cite{CAM} + $\mathcal{Y}$ &  & 59.4 & 62.1 & 60.4 & {60.6} &  & 63.6 & 59.1 & 67.4 & {63.4} &  & 58.3 & 63.2 & 58.5 & {60.0} && {60.1} \\
    \cline{1-1}\cline{3-6}\cline{8-11}\cline{13-16}\cline{18-18} \vspace{-1em}\\
    \oursem &  & 62.3 & 62.2 & 63.1 & {62.5} &  & 54.9 & 42.2 & {60.3} & {52.5} &  & {61.3} & {64.4} & {62.5} & {62.7} && {59.2}\\
    \oursem + $\mathcal{Y}$ &  & \textbf{62.8} & 62.3 & 63.4 &{62.8} &  & 64.8 & \textbf{60.3} & \textbf{71.0} & \textbf{65.4} &  & \textbf{61.3} & \textbf{64.4} & \textbf{62.5} & \textbf{62.7} && \textbf{63.6}\\
    \cline{1-1}\cline{3-6}\cline{8-11}\cline{13-16}\cline{18-18} \vspace{-1em}\\
    \end{tabular}
    
    \vspace{.5em}
    \caption{\small\textbf{WSOL results on CUB, OpenImages, and ImageNet.} Extension of Table~\textcolor{red}{2} in main paper. \oursem and \oursem+$\mathcal{Y}$ are compared against the baseline methods. \oursem+$\mathcal{Y}$ denote the aggregated attribution map for classes in $\mathcal{Y}$.}
    \label{appendix:tab:wsol-results}
\end{table*}



\end{document}